\documentclass[10pt,twocolumn,letterpaper]{article}

\usepackage[pagenumbers]{cvpr} 

\usepackage{blindtext}
\usepackage{booktabs}
\usepackage{amsmath} 
\usepackage{quiver}
\usepackage{graphicx}
\usepackage{multirow}
\usepackage[table]{xcolor}
\usepackage{amssymb}
\usepackage{pifont}
\usepackage{float}

\usepackage{pgfplots}
\pgfplotsset{compat=1.18}
\usepackage{pgfplotstable}
\usetikzlibrary{patterns}

\newcommand{\ours}{SPAR}

\definecolor{cvprblue}{rgb}{0.21,0.49,0.74}
\usepackage[pagebackref,breaklinks,colorlinks,allcolors=cvprblue]{hyperref}

\def\paperID{37526} 
\def\confName{CVPR}
\def\confYear{2026}

\title{SPAR: Single-Pass Any-Resolution ViT for Open-vocabulary Segmentation}

\author{
Naomi Kombol\textsuperscript{1}
\qquad
Ivan Martinović\textsuperscript{1} 
\qquad
Siniša Šegvić\textsuperscript{1} 
\qquad
Giorgos Tolias\textsuperscript{2} \\[0.52em]
\renewcommand{\arraystretch}{0.9}
\begin{tabular}{@{}cc@{}}
\textsuperscript{1}Faculty of Electrical Engineering and Computing &
\textsuperscript{2}VRG, Faculty of Electrical Engineering \\
University of Zagreb & Czech Technical University in Prague \\[-0.38em] 
\end{tabular}
}

\begin{document}
\maketitle
\vspace{-100pt}
\begin{abstract}
Foundational Vision Transformers (ViTs) have limited effectiveness in tasks requiring fine-grained spatial understanding, due to their fixed pre-training resolution and inherently coarse patch-level representations. These challenges are especially pronounced in dense prediction scenarios, such as open-vocabulary segmentation with ViT-based vision-language models, where high-resolution inputs are essential for accurate pixel-level reasoning. Existing approaches typically process large-resolution images using a sliding-window strategy at the pre-training resolution. While this improves accuracy through finer strides, it comes at a significant computational cost. We introduce \textbf{\ours{}}: \textbf{S}ingle-\textbf{P}ass \textbf{A}ny-\textbf{R}esolution ViT, a resolution-agnostic dense feature 
extractor
designed for efficient high-resolution inference. 
We distill the spatial reasoning capabilities of a finely-strided, sliding-window teacher into a single-pass student using a feature regression loss, without requiring architectural changes or pixel-level supervision. 
Applied to open-vocabulary segmentation, \ours{} improves single-pass baselines by up to 10.5 mIoU and even surpasses the teacher, demonstrating effectiveness in efficient, high-resolution reasoning. Code: \href{https://github.com/naomikombol/SPAR}{https://github.com/naomikombol/SPAR}
\end{abstract}    
\section{Introduction}
\label{sec:intro}

\begin{figure}[t]
  \vspace{-8pt}
  \centering
   \definecolor{GoogleBlue}{HTML}{2196F3}
\definecolor{GoogleRed}{HTML}{F44336}
\definecolor{GoogleGreen}{HTML}{4CAF50}
\definecolor{GoogleYellow}{HTML}{FFEB3B}
\begin{tikzpicture}
\begin{axis}[
    xlabel={\small inference time (s)},
    ylabel={\small avg. mIoU over 6 datasets}, 
    grid=both,
    major grid style={color=gray!50, opacity=0.7},
    minor grid style={color=gray!40, opacity=0.7},
    width=\columnwidth,
    height=6.cm,
    ymin=32,ymax=45.5,
    xmin=0,xmax=350,
    xmode=log,
    legend style={font=\footnotesize,row sep=-1pt,inner xsep=1pt,inner ysep=0pt},
    legend pos=south east,
    legend cell align={left},
]
\addlegendimage{only marks, mark=square*, mark options={fill=GoogleBlue, draw=GoogleBlue}, mark size=2.5pt} \addlegendentry{Sliding-window};
\addlegendimage{only marks, mark=square*, mark options={fill=GoogleYellow, draw=GoogleYellow}, mark size=2.5pt} \addlegendentry{Sliding-window, ND*};
\addlegendimage{only marks, mark=triangle*, mark options={fill=GoogleGreen, draw=GoogleGreen}, mark size=2.5pt} \addlegendentry{Single-pass};
\addlegendimage{only marks, mark=pentagon*, mark options={fill=GoogleRed, draw=GoogleRed}, mark size=2.5pt} \addlegendentry{\ours{}};

\addplot+[only marks, mark=triangle*, mark options={fill=GoogleGreen, draw=GoogleGreen}, mark size=4pt] coordinates {(0.47882,33.0)}; 

\addplot+[only marks, mark=square*, mark options={fill=GoogleYellow, draw=GoogleYellow}, mark size=2.5pt, forget plot] coordinates {(209.5953,41.5)}; 
\node at (axis cs:209.5953,41.6) [anchor=south] {\scriptsize 8};

\addplot+[only marks, mark=square*, mark options={fill=GoogleBlue, draw=GoogleBlue}, mark size=2.5pt, forget plot] coordinates {(53.1765,39.7)}; 
\node at (axis cs:63.1765,39.8) [anchor=south] {\scriptsize 16};
\addplot+[only marks, mark=square*, mark options={fill=GoogleBlue, draw=GoogleBlue}, mark size=2.5pt, forget plot] coordinates {(13.832,39.4)}; 
\node at (axis cs:13.832,39.5) [anchor=south] {\scriptsize 32};

\addplot+[only marks, mark=square*, mark options={fill=GoogleBlue, draw=GoogleBlue}, mark size=2.5pt, forget plot] coordinates {(3.77278,38.1)}; 
\node at (axis cs:3.77278,38.2) [anchor=south] {\scriptsize 64};
\addplot+[only marks, mark=square*, mark options={fill=GoogleBlue, draw=GoogleBlue}, mark size=2.5pt, forget plot] coordinates {(1.09164,37.1)}; 
\node at (axis cs:1.09164,37.2) [anchor=south] {\scriptsize 128};
\addplot+[only marks, mark=square*, mark options={fill=GoogleBlue, draw=GoogleBlue}, mark size=2.5pt, forget plot] coordinates {(0.34746,35.5)}; 
\node at (axis cs:0.34746,35.6) [anchor=south] {\scriptsize 256};    
\addplot+[only marks, mark=square*, mark options={fill=GoogleBlue, draw=GoogleBlue}, mark size=2.5pt, forget plot] coordinates {(0.13422,33.9)}; 
\node at (axis cs:0.13422,34) [anchor=south] {\scriptsize 512};

\addplot+[only marks, mark=square*, mark options={fill=GoogleYellow, draw=GoogleYellow}, mark size=2.5pt, forget plot] coordinates {(24.84658,41.2)}; 
\node at (axis cs:24.84658,41.3) [anchor=south] {\scriptsize 24};
\addplot+[only marks, mark=square*, mark options={fill=GoogleYellow, draw=GoogleYellow}, mark size=2.5pt, forget plot] coordinates {(9.29058,40.8)}; 
\node at (axis cs: 9.29058,40.9) [anchor=south] {\scriptsize 40};

\addplot+[only marks, mark=square*, mark options={fill=GoogleYellow, draw=GoogleYellow}, mark size=2.5pt, forget plot] coordinates {(5.29222,40.0)}; 
\node at (axis cs: 5.29222,40.1) [anchor=south] {\scriptsize 56};

\addplot+[only marks, mark=pentagon*, mark options={fill=GoogleRed, draw=GoogleRed}, mark size=4pt] coordinates {(0.47882,43.6)}; 

\draw[<->, thick,  GoogleGreen!40!black, >=Stealth, opacity=1]
  (axis cs:0.47382,33.8) -- (axis cs:0.47882,42.8)
  node[midway, right=1pt, font=\footnotesize,
       fill=white, fill opacity=1, inner sep=1pt, rounded corners=1pt,
       yshift=14pt, text=GoogleGreen!80!black]{\textbf{+10.5}};

\draw[<->, thick, GoogleGreen!40!black, >=Stealth]
  (axis cs:0.67882,43.5) -- (axis cs:19,41.9)
  node[midway, above, yshift=4pt, font=\footnotesize,
     fill=white, inner sep=1pt, opacity=1, rounded corners=1pt,
     text=GoogleGreen!80!black]{\textbf{52x faster}};

\end{axis}
\end{tikzpicture}
   \vspace{-10pt}
   \caption{\textbf{Performance \vs inference time trade-off.} Comparison between the pre-trained SigLIP2 -- \texttt{ViT-B-16} with single-pass or sliding-window (stride value reported in text labels) inference, and our single-pass \ours{}-distilled model. 
   The teacher has sliding windows of size $512\times512$ with stride 24. 
   We report average performance across six datasets along with average inference time for an 1024$\times$2048 image.  *ND: stride not divisible by patch size.
   \vspace{-20pt}
   \label{fig:stride_vs_performance}}
\end{figure}
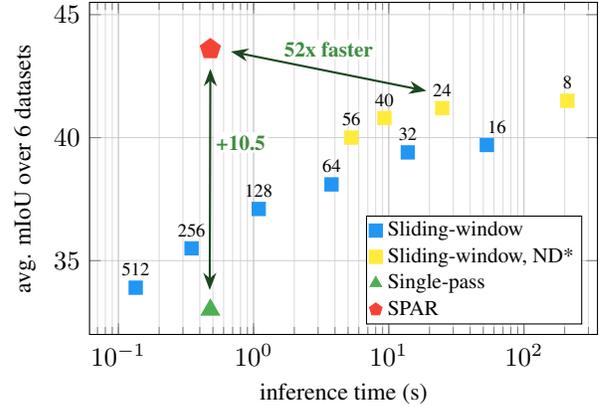

Vision Transformers (ViTs)~\citep{vit} have become the backbone of modern computer vision, powering foundation models like CLIP~\citep{CLIP}, DINO~\citep{dino, jose2025dinov2, dinov3}, and SigLIP~\citep{siglip, SigLIP2} through contrastive and self-supervised learning. ViTs process images as sequences of patches using self-attention~\citep{vaswani2017attention}, but due to its quadratic complexity, are typically pre-trained at a single, low resolution. While this suffices for image-level, it limits performance on dense prediction tasks like segmentation, which require fine-grained, per-pixel understanding.

This limitation is particularly severe in training-free open-vocabulary segmentation (OVS), where models must segment categories specified by text without access to pixel-level supervision. OVS approaches leverage ViTs as image feature extractors within Vision-Language models (VLMs), like CLIP~\citep{CLIP}, for their strong pre-training, but struggle with high-resolution test inputs due to discrepancies with the training resolution. Two common strategies attempt to address this issue: (i) interpolating positional encodings and (ii) sliding-window inference. 

The first approach supports single-pass processing and is efficient, but yields poor accuracy. In contrast, we show that the second approach, when using small strides and highly overlapping windows, significantly improves performance. Sliding-window inference enables a patch to appear in multiple contexts and enhances prediction accuracy through aggregation in overlapping regions. Interestingly, strides that are not divisible by the patch size lead to even better results, as sub-patch areas are exposed to diverse contexts.

However, sliding-window inference with small strides increases computational cost, making it impractical for real-world use (see \cref{fig:stride_vs_performance}). Despite growing architectural advances~\citep{Pix2Struct, NaVit, SigLIP2, dinov3} designed to handle variable resolutions, they are either not widely adopted or, as indicated in our experiments, have unexplored room for improvement on standard OVS benchmarks.

To address these challenges, we propose \ours{}: a method for enabling efficient resolution-agnostic feature extraction in ViTs via teacher-student distillation. \ours{} transfers the spatial reasoning of a frozen VLM teacher that operates in a finely-strided sliding-window manner, to a fast, single-pass student. The student learns to align its image features with the teacher’s using a regression loss, without requiring architectural changes or additional supervision. As shown in \cref{fig:stride_vs_performance}, \ours{} surpasses the teacher’s performance while maintaining single-pass efficiency.

SPAR uses the unchanged ViT architecture of the VLM and is compatible with a wide range of ViT backbones. During training, the student is exposed to diverse resolutions and aspect ratios, promoting robust generalization. We find that unfreezing only a small subset of layers suffices for strong resolution tolerance. To reduce compute, teacher features are precomputed and reused across training iterations. 
Our OVS experiments with vanilla zero-shot predictors show that \ours{} is effective with ViTs of OpenCLIP~\cite{openclip}, SigLIP2~\citep{SigLIP2}, but also DINOv3.txt~\citep{dinov3}, which already includes components making it resolution resilient.

In summary, we introduce SPAR, a framework for producing resolution-agnostic ViTs that deliver strong accuracy in a single forward pass. SPAR is applied to the vision encoder of common VLMs and requires no architectural modifications or additional labels. Rather than introducing new components, \ours{} focuses on a principled training strategy that ingrains resolution flexibility without sacrificing efficiency or feature-space alignment. The resulting model excels in training-free open-vocabulary segmentation across resolutions, matching or surpassing finely-strided sliding-window teachers while being up to $52\times$ faster.

\vspace{10pt}
\section{Related Work}
\label{sec:related_work}

\textbf{Open-vocabulary semantic segmentation (OVSS).}
Unlike standard semantic segmentation \citep{zheng2021setr, ZegFormer}, open-vocabulary semantic segmentation (OVSS) allows specifying and recognizing arbitrary categories using text. Progress on this task has closely followed advances in Vision–Language Models (VLMs). CLIP~\cite{CLIP} introduces contrastive pretraining on web-scale image–text pairs, demonstrating remarkable zero-shot classification capability. 
Subsequent models, such as ALIGN~\cite{ALIGN} and the SigLIP family~\cite{siglip,SigLIP2}, further scale datasets and refine training objectives -- the latter transitioning from contrastive formulations to more efficient, sigmoid-based matching losses -- thereby improving both semantic alignment and localization.
LSeg~\cite{LSeg} is the first to adapt CLIP for OVSS by aligning per-pixel visual features with frozen text embeddings. MaskCLIP~\cite{MaskCLIP} exposes spatial cues within CLIP’s vision tower by modifying its final attention layer to recover fine-grained localization.

Current methods are broadly categorized as \emph{training-based}~\cite{LSeg,OpenSeg,ZegFormer,catseg,jose2025dinov2} and \emph{training-free}~\cite{MaskCLIP,SCLIP,ProxyCLIP,CorrCLIP}.
Training-based approaches include: (i) methods that correlate text embeddings with regions produced by decoupled mask proposal generators~\cite{OpenSeg,ZegFormer,ZSEG,OVSeg,ODISE}; (ii) models that couple CLIP features with proposal generation~\cite{FCCLIP}; and (iii) variants that directly adapt CLIP for dense prediction~\cite{FCCLIP,MaskCLIP,ClearCLIP}.
Training-free methods, on the other hand, aim to preserve the pre-trained weights of CLIP without any fine-tuning. Some rectify localization weaknesses within CLIP itself by refining attention calculation and token intermixing ~\cite{MaskCLIP,SCLIP,ClearCLIP,ITACLIP}, others incorporate spatial priors from vision foundation models~\cite{ProxyCLIP,Trident,CorrCLIP}, or aid localization with generative priors~\cite{OVDiff,DiffSegmenter,CLIPer}.
Another line of work performs propagation of features or predictions to improve their spatial consistency~\cite{dinoiser,stojnic2025lposs}.
Across all these works, Vision Transformers (ViTs) remain the dominant backbone, and their ingrained resolution sensitivity persists as a shared limitation; one that our method addresses.

\noindent\textbf{Resolution-agnostic transformers.}
While Vision Transformers (ViTs) are powerful, fixed-resolution pretraining limits their robustness to varying input sizes. The naive remedy of interpolating positional encodings to match new image resolutions causes notable performance degradation~\cite{SwinTransformer,DeIT,CPE,ResFormer}. Pix2Struct~\cite{Pix2Struct} is among the first to address this issue by fixing token sequence length and resizing images with aspect ratios intact, breaking from traditional sliding-window inference~\cite{conv_window_slide,SwinTransformer}.
Building on this idea, several works introduce multi-resolution training to aid robustness. ResFormer~\cite{ResFormer} trains on square crops of varying sizes and expands positional encodings during attention with neighborhood-focused components. FlexiViT~\cite{FlexiViT} removes fixed patch sizes, randomly sampling them to expose the model to diverse sequence lengths. NaViT~\cite{NaVit} factorizes positional encodings along spatial axes, enabling processing of arbitrary-resolution images with native aspect ratios. SigLIP2's \textit{NaFlex}~\cite{SigLIP2}  unifies these ideas, combining flexible patching with variable sequence lengths for improved resolution generalization. However, NaFlex remains optimized for image–text alignment at the image- and not patch-level. This limits its use for dense prediction tasks like OVSS, as demonstrated in~\cref{sec:experiments}. Overall, these methods enhance robustness through diverse pretraining resolutions rather than post hoc adaptation.

Recent works continue this trend. ViTAR~\cite{ViTAR} expands training resolutions via fuzzy positional encoding, using lossy compression to reduce attention costs, while UniViTAR~\cite{UniViTAR} merges image and video modeling with aspect ratio preservation, introducing non-standard ViT changes. OryxMLLM~\cite{OryxMLLM} trains on native resolutions using dynamic feature compression.
In contrast, our approach adapts ViTs to varying resolutions \textit{post hoc}, without requiring additional labeled data or architectural modifications.
\looseness=-1

\section{Method}
\label{sec:method}

\begin{figure*}[t]
  \centering
  \includegraphics[width=0.93\textwidth]{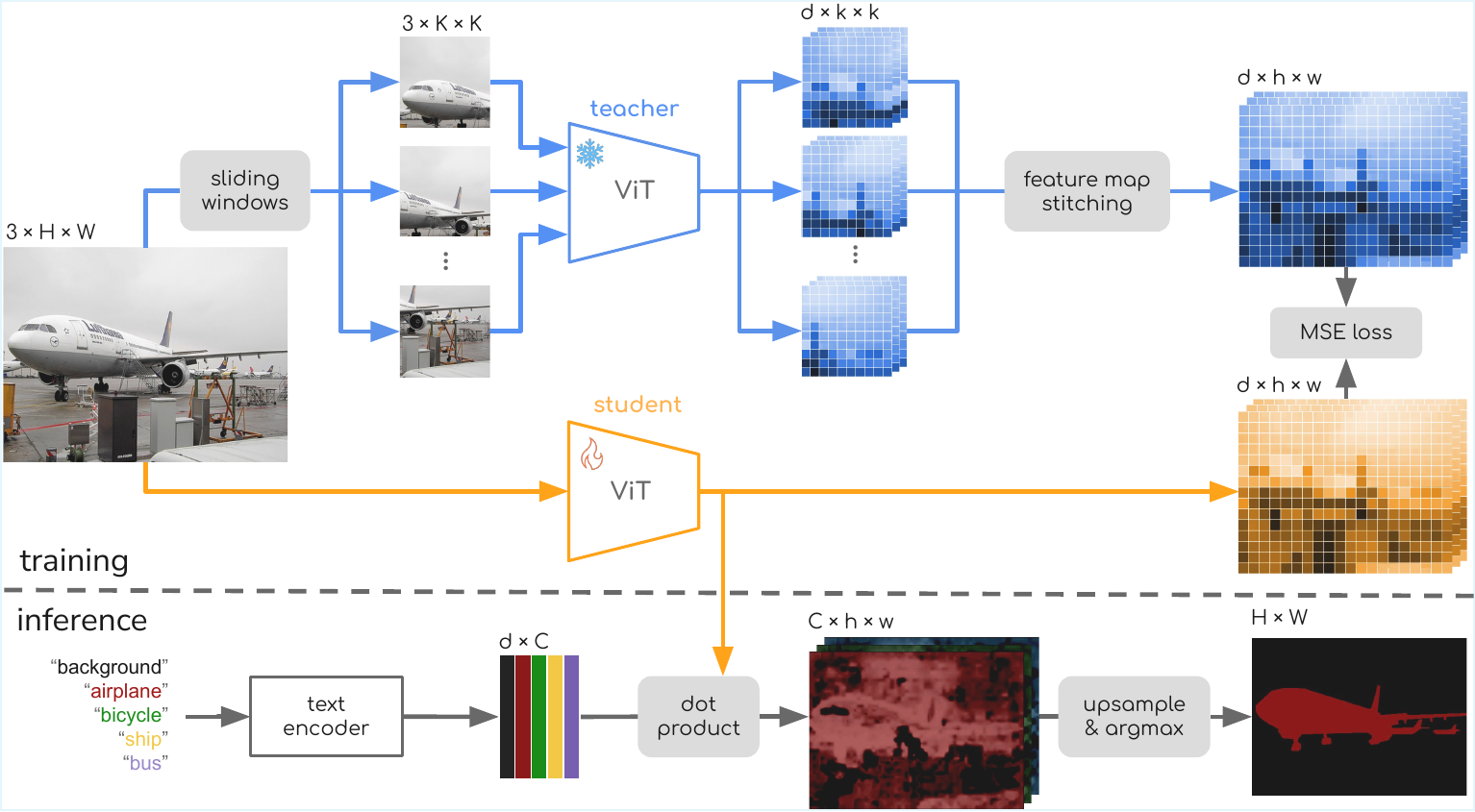}
  \vspace{-8pt}
\caption{\textbf{Overview of \ours{}.}
During training, the teacher branch uses a frozen foundational vision encoder to generate feature maps via a sliding-window process followed by stitching. Stitching refers to merging the feature maps of overlapping windows into a unified representation aligned with the original image layout. The student branch, initialized from the same pre-trained weights, trains to match the teacher's output using efficient single-pass inference. 
At inference time, the student model enables fast and accurate segmentation at diverse resolutions and aspect ratios using a single forward pass.
\vspace{-12pt}
  \label{fig:anyres}}
\end{figure*}

\subsection{Task Definition}
Open Vocabulary Segmentation (OVS) aims to assign a semantic class to every pixel in an image $X \in \mathbb{R}^{3 \times H \times W}$. The set of classes $\mathcal{C}$, containing $C$ class names, is arbitrary and can be specified at inference time.

\subsection{Preliminary}
Vision Transformers (ViTs) process images by dividing them into square patches, projecting each patch into a vector representation, and adding positional encodings based on patch location. These vectors are concatenated into a sequence and passed through transformer blocks. A special CLS token is typically included, but we omit it here.

Foundational ViTs are traditionally pre-trained on square images of fixed $K \times K$ image resolution and consist of an encoder $f : \mathbb{R}^{3 \times K \times K} \rightarrow \mathbb{R}^{d \times n}$ that produces a feature map $V = f(X) \in \mathbb{R}^{d \times n}$ per input image, where $n = k \cdot k$ is the number of patches, $d$ is the feature dimension, and $(k,k)$ are the spatial dimensions after reshaping into a tensor of size $d \times k \times k$. Due to patching, $k = K / P$, where $P$ is the patch size. Positional encodings are learned for that resolution only, \ie on a $k \times k$ grid of possible positions.

Vision-Language Models (VLMs) combine a vision encoder with a text encoder to map images and text into a shared feature space. Let $F \in \mathbb{R}^{d \times C}$ denote the concatenated text features for all classes in $\mathcal{C}$. 

OVS is performed by computing dot-product similarities between normalized image features and text features. The 2D map of class similarities for all classes given by
\[
Y(X) = \mathrm{norm}(V)^\top \mathrm{norm}(F) \in [0,1]^{C \times n},
\]
where $\mathrm{norm}(\cdot)$ denotes $\ell_2$ normalization applied along the feature dimension, is reshaped and upsampled to the resolution of $X$ using bilinear interpolation. Upsampling is necessary as the feature map $V$ has low resolution due to the patchification of input during inference, which limits segmentation quality, especially for small objects. Operating at higher resolutions, along with arbitrary aspect ratios, is critical for accurate segmentation.
Therefore, for image $X \in \mathbb{R}^{3 \times H \times W}$ of arbitrary size, we require an encoder capable of $f: \mathbb{R}^{3 \times H \times W} \rightarrow \mathbb{R}^{d \times n}$, with $n=h \cdot w$, that produces feature maps of spatial resolution proportional to the input size, \ie $h=H/P$, $w=W/P$. 

\subsection{Pre-trained Baselines}
\textbf{Single-pass} inference at arbitrary resolutions is feasible with ViT-based VLMs since self-attention can handle sequences of varying length. The required adjustment is interpolating positional encodings to match the new resolution. However, this approach suffers from performance degradation due to 
the training-inference resolution mismatch, and due to interpolation, which disrupts the absolute understanding of positional information learned during training.

\textbf{Sliding-window} is a common strategy for processing an arbitrary-resolution image $X \in \mathbb{R}^{3 \times H \times W}$. The image is divided into $m$ overlapping, or at least adjacent, windows of size $K \times K$, each processed independently. We denote the image windows by $\{ X_{w_i} \}_{i=1}^m$, where $X_{w_i} \in \mathbb{R}^{3 \times K \times K}$. The stride $s$ denotes how much we \emph{stride} between neighboring windows horizontally and/or vertically, controlling their overlap.

Feature maps are computed as $V_{w_i} = f(X_{w_i})$, and class similarities as $Y(X_{w_i}) = V_{w_i}^\top F$. The final prediction is obtained by stitching:
\begin{equation}
Y_\text{stitch}(X) = \text{stitch}(\{ Y(X_{w_i}) \}_{i=1}^m) \in [0,1]^{C \times h \times w}.
\end{equation}
Stitching refers to merging the individual window predictions into a single coherent output map, typically by averaging overlapping regions and aligning them to their original spatial positions.

This method preserves the native resolution of the vision encoder, avoiding single-pass issues, but incurs higher computational cost proportional to the total number of windows $m$. Even with batch processing, sliding-window inference is slower than single-pass for small strides. Nevertheless, small strides yield better performance by increasing window overlap, allowing each pixel to be seen in multiple contexts. This redundancy improves robustness through prediction averaging, akin to test-time augmentation.

\subsection{Single-Pass Any-Resolution (\ours{})}
\textbf{Performance \vs inference time trade-off.} Observing the trade-off between speed and accuracy in single-pass versus sliding-window inference (see \cref{fig:stride_vs_performance}), we propose a feature distillation approach that combines their advantages. A student model is trained to mimic the feature embeddings of a sliding-window teacher while maintaining the efficiency of single-pass inference. Both models have the same architecture, allowing the student to be initialized with the same weights and keep the same feature space.
A method overview is presented in Figure \ref{fig:anyres}.

\noindent\textbf{Sliding-window teacher.}  
Assume $X \in \mathbb{R}^{3 \times H \times W}$ is a training image of arbitrary resolution and aspect ratio. Instead of stitching predictions, we stitch features over windows of size $K \times K$ producing the teacher feature map by
\[
V_\text{teacher}(X) = \text{stitch}(\{ f(X_{w_i}) \}_{i=1}^m),
\]
where stitching refers to merging feature maps $f(X_{w_i})$ from $m$ windows into a single coherent representation aligned to the feature map layout that the single-pass model would produce, by averaging overlapping regions.

\noindent\textbf{Distillation loss.} 
The student model $g: \mathbb{R}^{3 \times H \times W} \rightarrow \mathbb{R}^{d \times n}$ processes the entire image in a single pass to produce $V_\text{student}(X) = g(X)$. The distillation loss minimizes the mean squared error between teacher and student features:
\[
\mathcal{L}_\text{distill} = \| V_\text{teacher}(X) - V_\text{student}(X) \|_2^2.
\]

\noindent\textbf{Training.}  
The teacher remains frozen during training.
We empirically observe that training only the last blocks of the student is a good choice for the standard OVS settings, but training all parameters excels at very large resolution inference and other tasks.
Optimization is performed over a dataset of images with diverse resolutions and aspect ratios. No annotations are required since the loss operates solely on feature maps; a generic image dataset suffices.

Training runs for multiple epochs, and teacher feature maps are precomputed and stored to save time. To enable easier feature-level stitching, all training images are resampled bilinearly to have dimensions divisible by the patch size $P$, and window size $K$ is also divisible by $P$. If stride $s$ is divisible by $P$, stitching occurs at the encoder’s native feature resolution due to full patch alignment. However, experiments show that using a stride not divisible by $P$ improves performance by exposing pixels to more diverse contexts. In such cases, if $s$ is divisible by $P/r$, stitching is performed by upsampling all feature maps by a factor of $r$ before merging, while then downsampling by a factor of $r$ after merging to restore the original feature resolution.

\definecolor{GoogleBlue}{HTML}{2196F3}
\definecolor{GoogleRed}{HTML}{F44336}
\definecolor{GoogleGreen}{HTML}{4CAF50}
\definecolor{GoogleYellow}{HTML}{FFEB3B}
\definecolor{GoogleGray}{HTML}{9AA0A6}
\definecolor{GoogleOrange}{HTML}{F88E16}
\definecolor{GoogleFuchsia}{HTML}{A00F50}

\section{Experiments}
\label{sec:experiments}
\subsection{Implementation Details}

\begin{table*}[t]
  \vspace{-8pt}
  \centering
    \newcommand{\cmark}{\ding{51}}
\newcommand{\xmark}{\ding{55}}
\newcommand*{\blarrow}{\rotatebox[origin=c]{270}{$\Rsh$}}
\definecolor{Improved}{HTML}{2CA02C} 
\newcommand{\g}[1]{\textcolor{gray}{#1}}
\def\arraystretch{0.92}
\setlength{\tabcolsep}{4.5pt}
\small
\begin{tabular}{lllcccccccll}
\toprule
     Models              & Inference   &  Upsampler      & Voc21 & Voc20 & CS & ADE & C60 & C59 & $\text{Mean}_{6}$ & &\\ 
    \midrule
    \rowcolor{gray!10} \multicolumn{12}{c}{SigLIP2\textsubscript{~\cite{SigLIP2}} -- \texttt{ViT-B-16}}\\
    NaFlex\textsubscript{~\cite{SigLIP2}}            & single-pass        & bilinear & 35.8 &66.6	&22.8	&16.2 &23.7	&25.2	&31.7 & & \\
     Pre-trained      &  sliding-window & bilinear   &45.0& 	77.5&\textbf{ 	 38.4}& 	21.8&	30.7&	34.0&	41.2 &  &\blarrow \\     
     Pre-trained      & single-pass        & bilinear   &36.1	&71.3 &23.5 &16.8 &24.5 &26.1 &33.1 & \blarrow&\\ 
     \ours{}          & single-pass        & bilinear& \textbf{47.3} & \textbf{81.5	}&\textbf{38.4} & \textbf{23.4}	& \textbf{33.8} & \textbf{37.2}	& \textbf{43.6} &  \textbf{\textcolor{Improved}{+10.5}} &\textbf{\textcolor{Improved}{+2.4}} \\
    \midrule
    LPOSS\textsubscript{~\cite{stojnic2025lposs}}+Pre-tr.    & single-pass        & bilinear& 46.1	&89.6	&34.5	&19.9	&30.9	&35.2	&42.7 & \blarrow & \\
     LPOSS\textsubscript{~\cite{stojnic2025lposs}}+\ours{}   & single-pass        & bilinear& \textbf{51.2	}&\textbf{89.7}	&\textbf{39.2	}&\textbf{25.8}	&\textbf{34.5}	&\textbf{39.8	}&\textbf{46.7} & \textbf{\textcolor{Improved}{+4.0}}&\\
    \midrule
     Pre-trained  & single-pass        & AnyUp\textsubscript{~\cite{AnyUp}} & 42.4 & 82.3 & 33.2 & 23.1 & 30.1 & 34.5 & 40.9 & \blarrow & \\ 
     \ours{}          & single-pass        & AnyUp\textsubscript{~\cite{AnyUp}} & \textbf{51.0} & \textbf{86.2} & \textbf{38.6} & \textbf{26.1} & \textbf{37.1} & \textbf{41.5} & \textbf{46.8} & \textbf{\textcolor{Improved}{+5.9}} & \\[5pt]
    \rowcolor{gray!10} \multicolumn{12}{c}{OpenCLIP\textsubscript{~\cite{openclip}} -- \texttt{ViT-B-16}}\\
     Pre-trained      & sliding-window & bilinear  &  \textbf{48.5} & 55.1 &  \textbf{30.8} & 15.4 & 25.2 & 26.6 & 33.6 & &\blarrow \\
     Pre-trained     & single-pass        & bilinear & 43.1 & 52.3 & 17.7 & 11.5 & 20.3 & 21.4 & 27.7  & \blarrow&\\
     \ours{}          & single-pass        & bilinear &  \textbf{48.5} &  \textbf{57.6} & 25.7 &  \textbf{16.2} &  \textbf{28.2} &  \textbf{30.3} &  \textbf{34.4} &  \textbf{\textcolor{Improved}{+6.7}}&\textbf{\textcolor{Improved}{+0.8}}\\  
     \midrule
    Pre-trained  & single-pass        & AnyUp\textsubscript{~\cite{AnyUp}} & 48.5 & \textbf{61.0} & 20.9 & 14.8 & 25.3 & 26.9 & 32.9 & \blarrow & \\
    \ours{}       & single-pass        & AnyUp\textsubscript{~\cite{AnyUp}} & \textbf{50.3} & 59.2 & \textbf{26.1} & \textbf{17.0} & \textbf{29.2} & \textbf{32.0}  & \textbf{35.6} &\textbf{\textcolor{Improved}{+2.7}} & \\[5pt]
    \rowcolor{gray!10} \multicolumn{12}{c}{DINOv3.txt\textsubscript{~\cite{dinov3}} -- \texttt{ViT-L-16}}\\
    Pre-trained       & sliding-window & bilinear & 42.6	&90.8	&39.7	&24.9	&\textbf{31.8	}&34.4	&44.0 & &\blarrow\\
    Pre-trained       & single-pass        & bilinear & \textbf{46.0}	&89.5	&35.9	&24.4	&32.1	&34.8	&43.8 & \blarrow & \\
    \ours{}           & single-pass        & bilinear & 43.1	&\textbf{91.3}	&\textbf{40.1}	&\textbf{25.4}	&31.6	&\textbf{35.0}	&\textbf{44.4} &\textbf{\textcolor{Improved}{+0.6}} & \textbf{\textcolor{Improved}{+0.4}}  \\ 
    \midrule
    Pre-trained   & single-pass         & AnyUp\textsubscript{~\cite{AnyUp}} & \textbf{46.2} & 89.9 & 32.4 & 24.9 & \textbf{32.4} & \textbf{35.1} & 43.5 & \blarrow& \\
    \ours{}       & single-pass         & AnyUp\textsubscript{~\cite{AnyUp}} & 42.8 & \textbf{91.5} & \textbf{36.1} & \textbf{25.6} & 31.7 & 35.0 & \textbf{43.8} & \textbf{\textcolor{Improved}{+0.3}} &\\
    \bottomrule
\end{tabular}

  \vspace{-8pt}
  \caption{\textbf{Performance evaluation via mIoU on six datasets.} We report results for 3 backbones comparing \ours{} to the pre-trained model and NaFlex \citep{SigLIP2}. Sliding-window inference uses stride $s=24$, which corresponds to the teacher model.
  We combine \ours{} with LPOSS \citep{stojnic2025lposs} to show complementarity with current open-vocabulary semantic segmentation methods.
  Prediction upsampling is either with bilinear interpolation or AnyUp \citep{AnyUp}. Labels as follows: 
  CS: Cityscapes, C60: Context60, C59: Context59.
  Best results per section are \textbf{bolded}.
  \label{tab:best_of_spar}}
\vspace{-10pt}
\end{table*}

\textbf{Models. }
We employ 3 pre-trained VLMs. 
Most experiments use SigLIP2 \citep{SigLIP2} with the \texttt{ViT-B-16} vision encoder, pre-trained on $512\times512$ images. SigLIP2 employs attention pooling over patch tokens with a learned query to obtain the final image-level representation. 
We skip the pooling step and directly project the value representation of patch tokens through the output linear layer of the attention-pooling and then the rest of the network. 
In this way, the patch representation we obtain is compatible with that of the text encoder. 
For the CLIP \citep{CLIP} experiments, we use OpenCLIP~\cite{openclip} with \texttt{ViT-B-16} and native image size $224\times224$. We follow MaskCLIP \citep{MaskCLIP} and set the last encoder's attention matrix to identity. 
Lastly, we use DINOv.3txt \citep{dinov3} with \texttt{ViT-L-16} and native image size $384\times384$. 
Inference is performed in accordance with the original instructions for semantic segmentation~\citep{dinov3}. 
As the image-level representation is a concatenation of a CLS token and the average of patch representations, we keep only the second half of the textual feature dimensions for later similarity comparison.

\noindent\textbf{Methods.} For \ours{}, we evaluate single-pass inference, and for pre-trained models, we use two variants. The \emph{single-pass baseline}, which has exactly the same runtime complexity as ours, and the \emph{sliding-window baseline} for different values of stride $s$. In particular, for $s=24$ this corresponds to the settings of the \emph{teacher model} during distillation. Note that, when evaluating sliding-window processing, stitching happens for class similarities at the pixel-level, while during distillation we stitch together upsampled vision features and resample back down to patch-level as discussed in~\ref{sec:method}. Additionally,  the $s=256$ case for SigLIP2, corresponds to the commonly used setting in OVS literature~\cite{stojnic2025lposs,dinoiser, SCLIP} of taking half the window size due to its low runtime cost. 
The window size $K$ is set to the native image size of each model, \ie $K=512$ for SigLIP2, $K=224$ for MaskCLIP and $K=384$ for DINOv3.txt.
We further show complementarity by combining \ours{} with learnable upsampling by AnyUp~\cite{AnyUp}, and label propagation performed by LPOSS~\cite{stojnic2025lposs}.
The former is a feature-agnostic model that operates independently per-feature dimension. Even though it is not its original use, we apply it to upsample class similarities and not the features themselves in all benchmarks, which has reduced complexity, \ie $C \ll D$. 
LPOSS is a training‑free OVSS approach that refines initial patch‑based predictions from a vision‑language model by propagating VLM-labels (CLIP) across patches in accordance with spatial and DINOv2-semantic similarity. For LPOSS, we replace both models with \ours{}-SigLIP2 to highlight its improved spatial coherence.
We additionally report NaFlex performance: SigLIP2's native variant to handle images of any resolution. 
Unless otherwise stated, bilinear interpolation is used to upsample predictions.

\begin{figure*}[t]
  \vspace{-3pt}
  \centering
    \begin{center}
    \ref{combinedlegend}
    \end{center}
  \vspace{-4pt}  
  \begin{tabular}{c@{\hspace{-10pt}}c}
    \pgfplotstableread{
resolution  gt     baseline  ours   wst_256 wst_24 naflex ours_all ours_all_ext
512         90.6   22.1      30.29  -       -      23.1   29.2     29.6
1024        93.6   22.8      36.03  27.63   35.47  23.5   35.2     36.8
1536        95.1   23.0      37.50  29.93   38.84  23.3   38.4     40.9
2048        95.1   21.3      34.52  30.44	39.45  22.4   38.4     41.14
3072        0      17.5      27.16  28.45   36.67  -      35.16    37.35
}{\perf}

\definecolor{GoogleBlue}{HTML}{2196F3}
\definecolor{GoogleRed}{HTML}{F44336}
\definecolor{GoogleGreen}{HTML}{4CAF50}
\definecolor{GoogleYellow}{HTML}{FFEB3B}
\definecolor{GoogleGray}{HTML}{9AA0A6}
\definecolor{GoogleOrange}{HTML}{F88E16}
\definecolor{GoogleFuchsia}{HTML}{A00F50}

\begin{tikzpicture}
\begin{axis}[
    width=0.48\linewidth,
    grid=both,
    major grid style={color=gray!50, opacity=0.7},
    minor grid style={color=gray!40, opacity=0.7},
    height=0.31\linewidth,
    ylabel={performance (mIoU)},
    xlabel={effective input image resolution (pixels)},
    legend cell align={left},
    legend pos=north west,
    legend style={
        cells={anchor=east},
        font=\footnotesize,
        row sep=-2.5pt,
        fill=white,          
        fill opacity=0.8,    
        draw=none,            
    },
    xtick={512,1024,1536,2048,3072},
    xticklabels={
        $512^2$,
        $1024^2$,
        $1536^2$,
        $2048^2$,
        $3072^2$
    },
]

\addplot[color=GoogleGreen, solid, mark=square*, mark size=2, line width=1.0, restrict x to domain=1024:3072]
    table[x=resolution, y expr={\thisrow{wst_256}}] \perf;

\addplot[color=GoogleBlue, solid, mark=square*, mark size=2, line width=1.0, restrict x to domain=1024:3072]
    table[x=resolution, y expr={\thisrow{wst_24}}] \perf;

\addplot[color=GoogleGray, solid, mark=*, mark size=1.5, line width=1.0, restrict x to domain=512:2048]
    table[x=resolution, y expr={\thisrow{naflex}}] \perf;

\addplot[color=GoogleYellow, solid, mark=triangle*, mark size=3, line width=1.0]
    table[x=resolution, y expr={\thisrow{baseline}}] \perf;

\addplot[color=GoogleRed, solid, mark=pentagon*, mark size=3, line width=1.0]
    table[x=resolution, y expr={\thisrow{ours}}] \perf;

\addplot[color=GoogleOrange, solid,  mark=pentagon*, mark size=3, line width=1.0, restrict x to domain=512:3072]
    table[x=resolution, y expr={\thisrow{ours_all}}] \perf;
    
\addplot[color=GoogleFuchsia, solid, mark=pentagon*, mark size=3, line width=1.0, line width=1.0, restrict x to domain=512:3072]
    table[x=resolution, y expr={\thisrow{ours_all_ext}}] \perf;

\end{axis}
\end{tikzpicture} & \pgfplotstableread{
resolution  single_pass  wst_256   wst_24
512         0.0251           -           - 
1024        0.1432      0.1667      6.7048
1536        0.61032     0.53494     29.69992
2048        1.862325    0.91618     67.8558
2560        4.53488     1.63844     123.008125
3072        9.1962      2.15573     191.6293
}{\perf}

\definecolor{GoogleBlue}{HTML}{2196F3}
\definecolor{GoogleRed}{HTML}{F44336}
\definecolor{GoogleGreen}{HTML}{4CAF50}
\definecolor{GoogleYellow}{HTML}{FFEB3B}
\definecolor{GoogleGray}{HTML}{9AA0A6}
\definecolor{GoogleOrange}{HTML}{F88E16}
\definecolor{GoogleFuchsia}{HTML}{A00F50}

\begin{tikzpicture}
\begin{axis}[
	width=0.48\linewidth,
	height=0.31\linewidth,
	ylabel={inference time (s)},
	xlabel={effective input image resolution (pixels)},
	grid=both,
    major grid style={color=gray!50, opacity=0.7},
    minor grid style={color=gray!40, opacity=0.7},
	ymode=log,
	legend to name=combinedlegend,   
	legend columns=-1,               
	legend style={
		draw=black,                
		column sep=6pt,            
		text=black,     
		inner sep=4pt,
		font=\small,
		/ tikz / every even column/.append style={column sep=10pt},
	},
    xtick={512,1024,1536,2048,3072},
    xticklabels={
        $512^2$,
        $1024^2$,
        $1536^2$,
        $2048^2$,
        $3072^2$
    },
]

\addlegendimage{only marks, color=GoogleGreen, mark=square*, mark size=2, line width=1.0};
\addlegendentry{Sliding-window, $s=256$};
\addlegendimage{only marks, color=GoogleBlue, mark=square*, mark size=2, line width=1.0};
\addlegendentry{Sliding-window, $s=24$};
\addlegendimage{only marks, color=GoogleGray, mark=*, mark size=1.5, line width=1.0};
\addlegendentry{NaFlex};
\addlegendimage{only marks, color=GoogleYellow, mark=triangle*, mark size=3, line width=1.0};
\addlegendentry{Single-pass};
\addlegendimage{only marks, color=GoogleRed, mark=pentagon*, mark size=3, line width=1.0};
\addlegendentry{\ours};

\addlegendimage{only marks, color=GoogleOrange, mark=pentagon*, mark size=3, line width=1.0};
\addlegendentry{\ours{} ALL};
\addlegendimage{only marks, color=GoogleFuchsia, mark=pentagon*, mark size=3, line width=1.0};
\addlegendentry{\ours{} ALL $\dagger$};

\addplot[color=GoogleGreen, solid, mark=square*, mark size=2, line width=1.0, restrict x to domain=1024:3072]
	table[x=resolution, y expr={\thisrow{wst_256}}]{\perf};
\addplot[color=GoogleBlue, solid, mark=square*, mark size=2, line width=1.0, restrict x to domain=1024:3072]
	table[x=resolution, y expr={\thisrow{wst_24}}]{\perf};
\addplot[color=GoogleFuchsia, solid, mark=pentagon*, mark size=3, line width=1.0, restrict x to domain=512:3072]
	table[x=resolution, y expr={\thisrow{single_pass}}]{\perf};

\end{axis}
\end{tikzpicture}
  \end{tabular}
\vspace{-10pt}  
  \caption{\textbf{Performance (left) and inference time (right) \vs image resolution.}
  We compare \ours{} to the pre-trained single-pass and sliding-window models, and single-pass NaFlex.
  SigLIP2 -- \texttt{ViT-B-16} with native resolution $512\times512$, and $K=512$ used. 
  Results on Cityscapes with resolution corresponding to the denoted area. Resolution 512 does not qualify for $K=512$. Single-pass models, including \ours{}, have the same inference time. 
  $\dagger$: larger resolution used in training (short side up to 2560 pixels instead of 2048). ALL: training the whole network, otherwise only the last 2 blocks.
  \label{fig:perf_reso}}
\vspace{-0pt}  
\end{figure*}
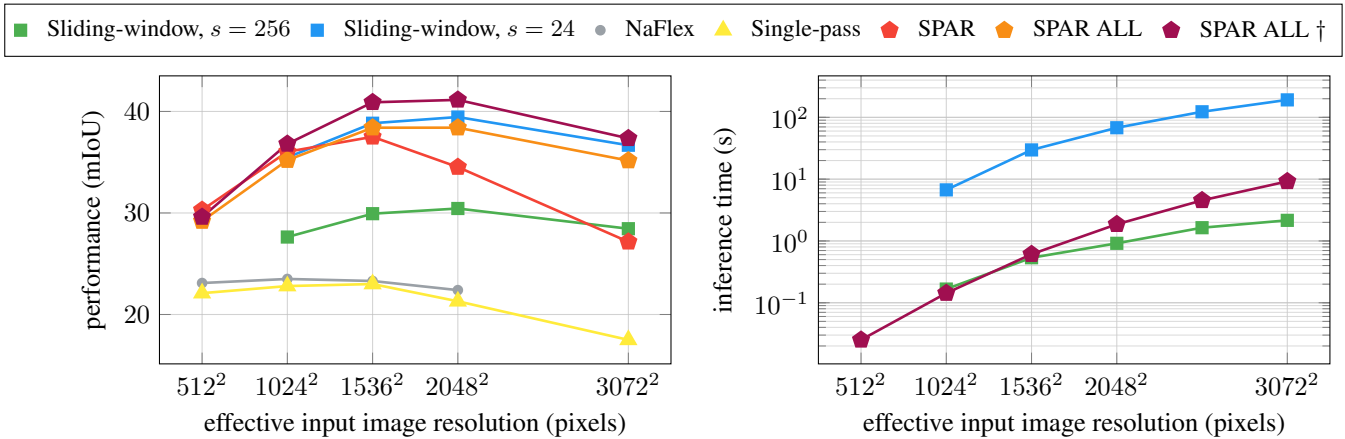

\noindent\textbf{Datasets and evaluation.}
We evaluate on six standard open-vocabulary semantic segmentation benchmarks derived from four datasets: Pascal VOC \citep{voc}, Pascal Context \citep{context}, ADE20K \citep{ade20k} and Cityscapes \citep{cityscapes}. VOC21 and Context60 variants include an additional background class over VOC20 and Context59. 
To allow evaluation for models that have large native resolution, \eg SigLIP2, we bilinearly resize images of all datasets to have their shorter side equal to 672, with the exception of Cityscapes, which is used at its original resolution.
We use mean Intersection over Union (mIoU) as the metric of choice for semantic segmentation.
By $\text{Mean}_{6}$ we denote the average mIoU over all six datasets. We train on a subset of 25k images from SA-1B~\cite{sam}, an 11-million-image dataset for class-agnostic semantic segmentation. Please note that we do not use any ground-truth annotations from SA-1B.

\noindent\textbf{Training of \ours{}.} 
For training, we chose the teacher with sliding windows and stride $s=24$ for its good balance of performance and inference time (see \cref{fig:stride_vs_performance}). 
We apply image augmentations once per image, as the stitched feature map from the teacher network is computed once and stored.
An image is augmented with random resizing of the shorter side to $[512-2048]$ in the default case, and $[512-2560]$ in the experiments with larger resolution.
We apply axis-independent random cropping, with side lengths from 512 to the maximum possible, and horizontal flipping, each with a 0.5 probability.

We train for 10 epochs with a constant learning rate $2 \cdot 10^{-5}$ with the AdamW \citep{adamw} optimizer and weight decay $10^{-4}$ while tuning only the ViT's last two blocks, unless otherwise mentioned. The default choice for the training set consists of 25k images from SA-1B~\citep{sam}. More training details in the supplement.

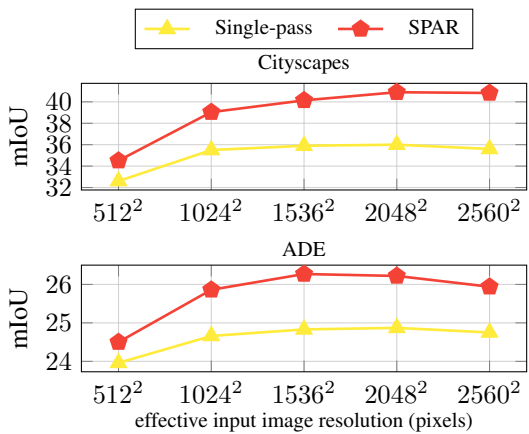
\begin{figure}[t]
  \centering
  \pgfplotstableread{
resolution  gt     baseline  dino_cs  spar_cs   dino_ade  spar_ade 
512         0       0        32.6  34.51       23.96       24.5 
1024        93.6   22.8      35.5   39.04      24.66       25.86
1536        95.1   23.0      35.9   40.14      24.83       26.27
2048        95.1   21.3      36.0   40.9       24.87       26.22
2560        0      18.92     35.6   40.83      24.75       25.94
}{\perf}

\definecolor{GoogleRed}{HTML}{F44336}
\definecolor{GoogleYellow}{HTML}{FFEB3B}
\definecolor{GoogleGreen}{HTML}{4CAF50}
\definecolor{GoogleBlue}{HTML}{2196F3}
\definecolor{GoogleGray}{HTML}{9AA0A6}

\begin{tikzpicture}

\begin{axis}[
    name=csaxis,
    width=0.9\linewidth,
    height=0.36\linewidth,
    title={\footnotesize Cityscapes},
    title style={yshift=-6pt}, 
    ylabel={mIoU},
    grid=both,
    major grid style={color=gray!50, opacity=0.7},
    minor grid style={color=gray!40, opacity=0.7},
    xtick={512,1024,1536,2048,2560},
    xticklabels={$512^2$,$1024^2$,$1536^2$,$2048^2$,$2560^2$},
    legend columns=2,
    legend style={
    at={(0.5,1.32)},   
    anchor=south,
    cells={anchor=west},
    font=\footnotesize,
    fill=white,
    fill opacity=1.0,
    draw=black,
    column sep=0.25cm
},
]

\addplot[color=GoogleYellow, solid, mark=triangle*, mark size=3, line width=1.0]
    table[x=resolution, y=dino_cs] \perf;
\addlegendentry{Single-pass}

\addplot[color=GoogleRed, solid, mark=pentagon*, mark size=3, line width=1.0]
    table[x=resolution, y=spar_cs] \perf;
\addlegendentry{\ours{}}

\end{axis}

\begin{axis}[
    at={(csaxis.south)},
    anchor=north,
    yshift=-1.0cm,  
    width=0.9\linewidth,
    height=0.36\linewidth,
    title={\footnotesize ADE},
    title style={yshift=-6pt},
    ylabel={mIoU},
    grid=both,
    major grid style={color=gray!50, opacity=0.7},
    minor grid style={color=gray!40, opacity=0.7},
    xtick={512,1024,1536,2048,2560},
    xticklabels={$512^2$,$1024^2$,$1536^2$,$2048^2$,$2560^2$},
    xlabel={\footnotesize effective input image resolution (pixels)},
    xlabel style={yshift=3pt},
]

\addplot[color=GoogleYellow, solid, mark=triangle*, mark size=3, line width=1.0]
    table[x=resolution, y=dino_ade] \perf;

\addplot[color=GoogleRed, solid, mark=pentagon*, mark size=3, line width=1.0]
    table[x=resolution, y=spar_ade] \perf;

\end{axis}

\end{tikzpicture}
\vspace{-10pt}  
  \caption{\textbf{Performance vs. resolution for \texttt{DINOv3.txt}.} } 
  \label{fig:timer_reso_perf_dino}
\vspace{-10pt}  
\end{figure}

\subsection{Main Results}

\textbf{\ours{} improvements and compatibility.}
We report the quantitative results across three different backbones 
in \cref{tab:best_of_spar}. 
Compared to pre-trained single-pass inference, \ours{} yields major gains of +10.5 and +6.7 mIoU for SigLIP2 and OpenCLIP, respectively. 
Moreover, \ours{} consistently surpasses the teacher across all models in average performance as well as individually on most datasets.

For DINOv3.txt, \ours{} produces only slight gains over the teacher. 
DINOv3.txt already performs well in single-pass inference, leaving limited room for distillation from sliding-windows.
This is likely due to the model's RoPE encodings~\cite{RoPE} and an additional high-res fine-tuning stage. 
Nevertheless, \ours{} on Cityscapes, which has larger resolution test images, achieves a substantial gain from 35.9 to 40.1 mIoU, effectively recovering performance on images that deviate most from the training resolution.

Moreover, we see that the SigLIP2 NaFlex variant does not perform competitively compared to either \ours{} or the sliding-window baseline, demonstrating its limitations for dense prediction tasks despite its additional training for resolution robustness.
AnyUp~\citep{AnyUp}, combined with the single-pass baseline, overall does not manage to exceed \ours{}. However, in tandem, it proves compatible and boosts the performance by an appreciable margin, notably by 3.2 on average for \ours{}-SigLIP2. Interestingly, AnyUp is not effective when combined with DINOv3.txt. 
Finally, the combination of \ours{} and LPOSS~\citep{stojnic2025lposs} further boosts performance (another +3.1 mIoU), demonstrating how \ours{}-trained models are complementary to other OVS methods.

\noindent\textbf{Performance on large resolution inference.}
Pre-trained single-pass inference and NaFlex do not benefit from larger resolutions, which is not the case for pre-trained sliding-window inference. 
The default variant of \ours{} benefits from increasing effective resolution up to $1.5$k$^2$, and surpasses the sliding-window approach when $s=256$ but not the more costly $s=24$. 
Nevertheless, we observe that training the full network improves \ours{}'s ability to perform large resolution inference, which is further enhanced by increasing the maximum shorter side length in training from 2048 to 2560. 
These two choices help with large, but not with small, resolution inference. 
Overall, \ours{} achieves the best performance across all inference-time resolutions.

Due to the observation in \cref{tab:best_of_spar} that with \ours{}, DINOv3.txt benefits at larger resolution, we replicate the same experiment for DINOv3.txt in \cref{fig:timer_reso_perf_dino}. We observe that \ours{} consistently improves DINOv3.txt performance across all resolutions, despite the design and training choices of DINOv3.txt that already promote resolution robustness.

\noindent\textbf{Performance \vs inference time.}
In \cref{fig:stride_vs_performance}, we analyze the trade-off between inference time and performance of \ours{} SigLIP2 and the pre-trained models for different variants.
\ours{} achieves higher results than the best sliding-window variant, while being $52\times$ faster than the teacher, and having a $+10.5$ point mIoU performance gain over the pre-trained single-pass model, while keeping the same inference time.
For the sliding-window variants, we observe smaller stride values leading to much better performance than the setting of $s=256$, \ie half the window size which is commonly adopted in the OVS literature due to its low runtime cost. 
Furthermore, without compromising inference time, we see that a stride value not divisible by patch size gives a good performance boost, justifying its use for the \ours{} teacher.

We further analyze the performance and inference time trade-off for increasing image resolution in \cref{fig:perf_reso}, with images scaled to the same area as denoted on the axis. 
Compared to single-pass inference, the sliding-window approach with $s=24$ incurs a substantial runtime cost, being two orders of magnitude slower. All \ours{} variants are noticeably cheaper, providing a great performance-cost trade-off.

\begin{table}[t]
  \centering
  \footnotesize
\setlength{\tabcolsep}{3.5pt}
\begin{tabular}{lr@{\hspace{10pt}}ccc}
\toprule
     Training Configuration &  \#params (M) & CS & ADE & $\text{Mean}_6$ \\ \midrule
    \rowcolor{gray!10} \multicolumn{5}{c}{Pre-trained model}\\
     Single-pass &  &23.5 &16.8  &33.1 \\ 
     Sliding-windows $s=24$ && 38.4	& 	21.8&	41.2\\   
    \rowcolor{gray!10} \multicolumn{5}{c}{\ours{} model}\\
     ALL params & 93.5 & \underline{38.5} &22.8	&42.5\\ 
     ALL params $\dagger$ & 93.5 &  \textbf{40.5	}&22.7 	&42.0 \\ 
     Last block & 7.1 &36.9	&\underline{23.3} & \underline{43.3} \\ 
     \textbf{Last 2 blocks (default)}  & 14.2 &38.4	&\textbf{23.4	}&\textbf{43.6}\\
     Last 2 blocks $\dagger$  & 14.2 & 38.5	&23.0 & 42.9 \\
     Last 3 blocks& 21.3 &37.5	&23.1	&42.9\\ 
     Patch projection & 0.6 &25.6	&21.4	&38.0 \\ 
     Positional encoding & 0.8&31.4	&21.4		&39.6  \\ 
     Last 2 blocks - \texttt{MLP} & 9.4&37.5	&22.6	&42.7 \\ 
     Last 2 blocks - \texttt{QKV} & 3.5 &34.9&22.2		&41.7 \\ 
    \bottomrule
\end{tabular}

\vspace{-8pt}  
  \caption{\textbf{Effect of fine-tuning different network elements}. We show mIoU on Cityscapes (CS), ADE20K, and the mean over six datasets for \ours{}-SigLIP2 -- \texttt{ViT-B-16}. Fine-tuning only the last two transformer blocks, \textbf{bolded} for emphasis, yields the best overall performance. \#params: number of fine-tuned parameters. $\dagger$: larger resolution used in training. Best result is \textbf{bolded}, second best is \underline{underlined}.}
  \label{tab:config_ablations_stride_24_teacher}
\vspace{-0pt}  
\end{table}

\begin{table}[t]
  \centering
  \small
    \footnotesize
\begin{tabular}{lccc}
\toprule
    
    Training Set & CS & ADE  & $\text{Mean}_6$\\ 
    \midrule
    \rowcolor{gray!10} \multicolumn{4}{c}{Pre-trained model}\\
    Sliding-window $s=24$& \underline{38.4}& 	21.8&	41.2 \\ 
    \rowcolor{gray!10}  \multicolumn{4}{c}{\ours{} model}\\
     ADE20k+CS+VOC  & \textbf{39.3} &\textbf{23.6}	&\underline{43.4}\\
     ADE20k & 37.4	&\underline{23.5}	&43.0\\[5pt]
     SA-1B ~1.25k (5\%)  & 35.4	&21.9	&41.1 \\ 
     SA-1B ~2.5k (10\%) &36.4	&22.4	&42.1 \\ 
     SA-1B 12.5k (50\%) & 38.0	&23.2	&43.2 \\ 
     SA-1B ~~25k  (100\%)&\underline{38.4	}&23.4	&\textbf{43.6}\\
     SA-1B ~~25k (diff. subset)  & 37.7	&23.2	&\textbf{43.6} \\ 
     SA-1B ~~50k  (200\%)& 38.1	&\underline{23.5	}&\textbf{43.6}\\
    
    \bottomrule
\end{tabular}

\vspace{-8pt}  
  \caption{\textbf{Impact of distillation data composition and scale}. We show mIoU on Cityscapes (CS), ADE20K, and the mean over six datasets for \ours{}-SigLIP2 -- \texttt{ViT-B-16} when training on different unlabeled sources. Comparable results across all settings show that \ours{} does not rely on in-domain data, and performance saturates with 25k distillation images. Best result is \textbf{bolded}, second best is \underline{underlined}.}
  \label{tab:siglip2_training_set_ablation}
\vspace{-0pt}  
\end{table}

\subsection{Additional Analyses}
\label{subsec:additional_analisys}

\textbf{Fine-tuning different network layers.} 
In \cref{tab:config_ablations_stride_24_teacher} we summarize the results of training assorted configurations of unfrozen parameters for \ours{} built on SigLIP2, reporting Cityscapes, ADE20K and mean mIoU over six datasets.
When varying the number of trainable blocks, fine-tuning only the final block already recovers, and slightly surpasses, the teacher’s performance, tuning the last two blocks yields a minor extra gain, while tuning more layers slightly degrades performance. 
To isolate the contribution of specific components, we fine-tune either only the \texttt{MLP} or the \texttt{QKV} projections. They  both perform worse than full block training, with 
\texttt{MLP} being slightly better than \texttt{QKV} by using far more parameters.
Tuning only the patch projection yields only a minor improvement compared to the single-pass pre-trained model, while training positional encodings achieves noticeably better performance, yet still not enough to surpass the sliding-window teacher.
Training all parameters or with larger resolutions does not immediately surpass the normal regime: their value is better displayed in the experiment with larger resolution inference (see \cref{fig:perf_reso}).

\begin{figure*}[h!]
\vspace{-0pt}
\centering
\setlength{\tabcolsep}{1pt}
\renewcommand{\arraystretch}{0.9}
\newcommand{\myheight}{0.135\linewidth}
\begin{tabular}{ccccccc}    

    \fontsize{7pt}{9.10pt}\selectfont  Image~(GT)  & 
    \fontsize{7pt}{9.10pt}\selectfont Single-pass  & 
    \fontsize{7pt}{9.10pt}\selectfont  Sliding-window & 
    \fontsize{7pt}{9.10pt}\selectfont \ours{} & 
    \fontsize{7pt}{9.10pt}\selectfont \ours{} + LPOSS  &
    \fontsize{7pt}{9.10pt}\selectfont Sliding-window PCA & 
    \fontsize{7pt}{9.10pt}\selectfont \ours{} PCA \\

    \includegraphics[width=\myheight]{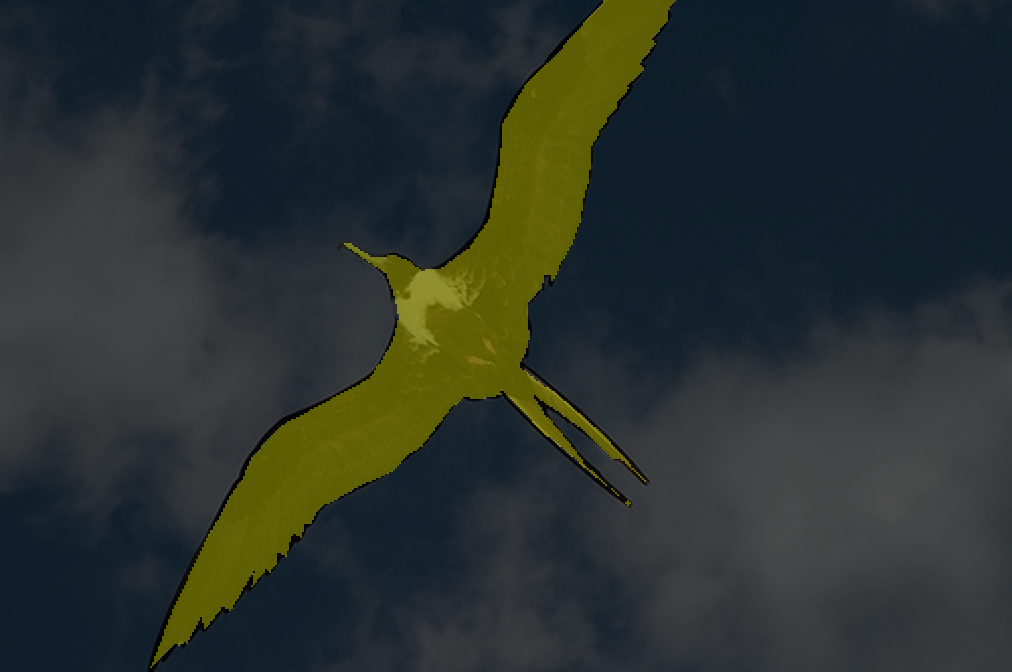} &
    \includegraphics[width=\myheight]{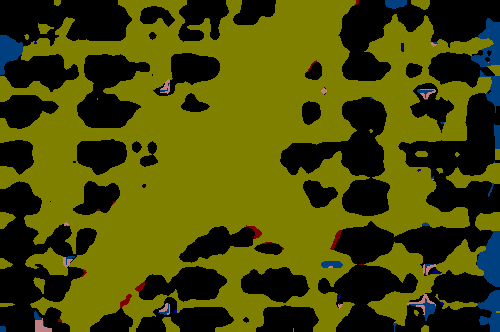} &
    \includegraphics[width=\myheight]{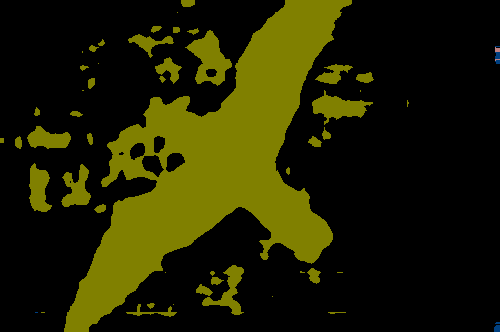} &
    \includegraphics[width=\myheight]{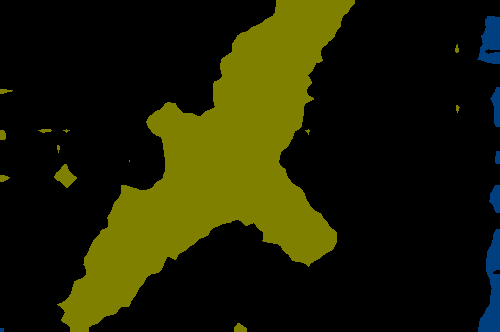} &
    \includegraphics[width=\myheight]{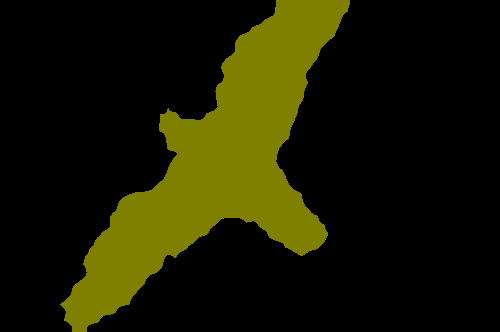}  &
    \includegraphics[width=\myheight]{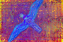}  &
    \includegraphics[width=\myheight]{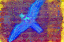}  \\

    \includegraphics[width=\myheight]{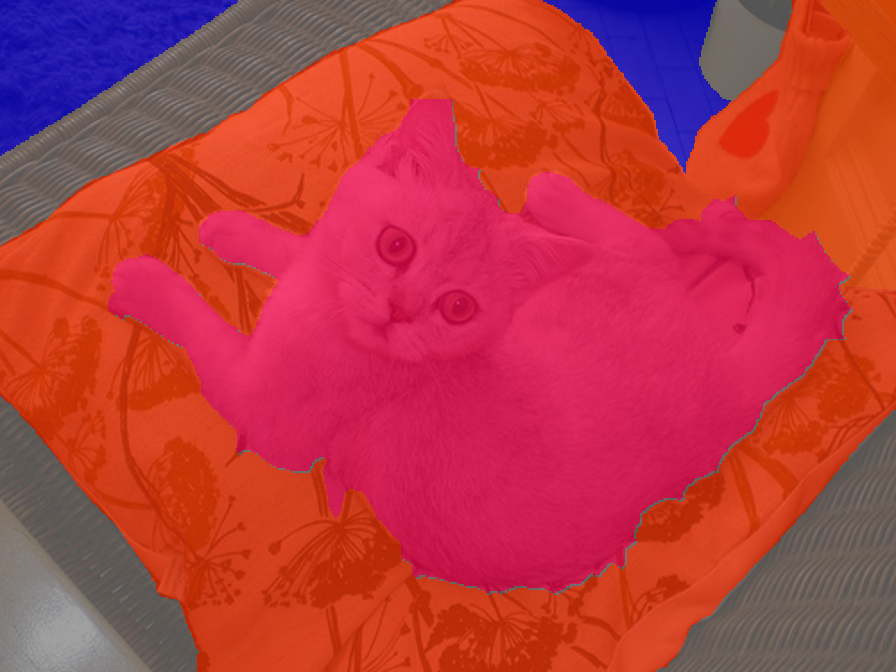} &
    \includegraphics[width=\myheight]{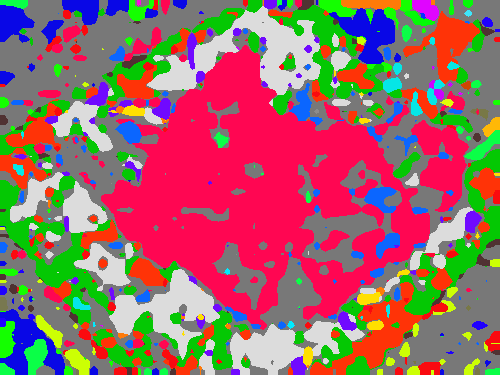} &
    \includegraphics[width=\myheight]{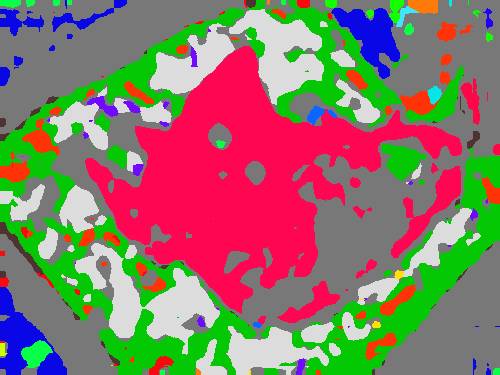} &
    \includegraphics[width=\myheight]{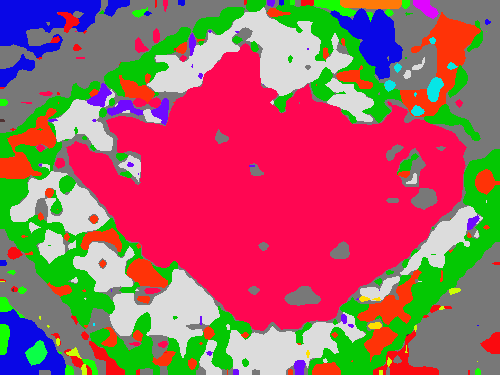} &
    \includegraphics[width=\myheight]{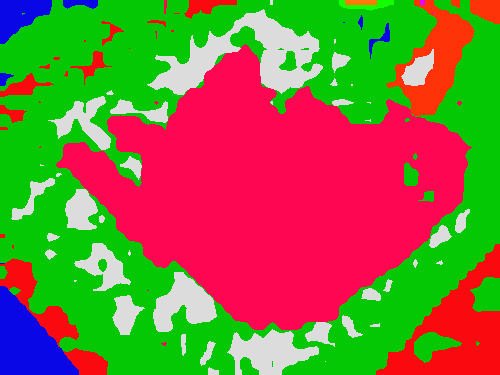} &
    \includegraphics[width=\myheight]{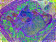}  &
    \includegraphics[width=\myheight]{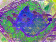}  \\

    \includegraphics[width=\myheight]{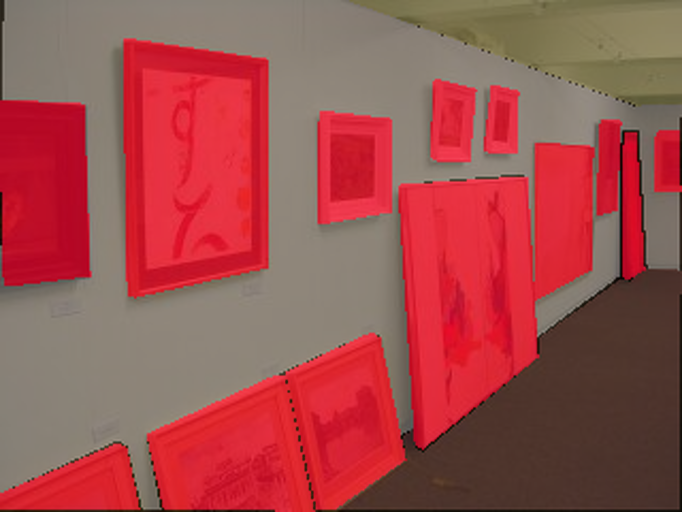} &
    \includegraphics[width=\myheight]{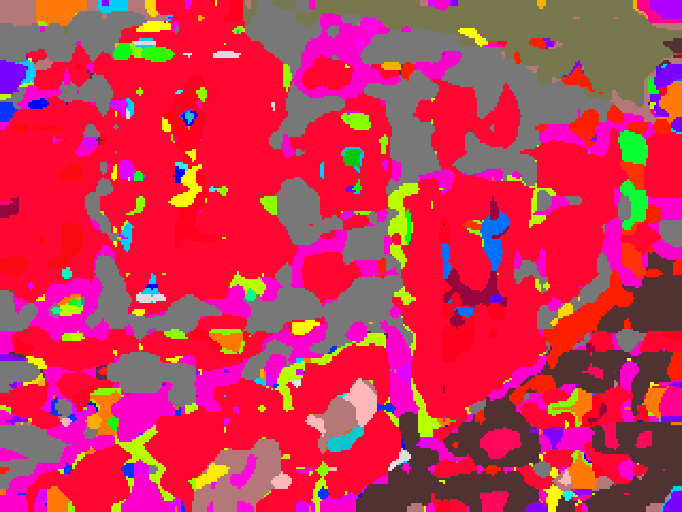} &
    \includegraphics[width=\myheight]{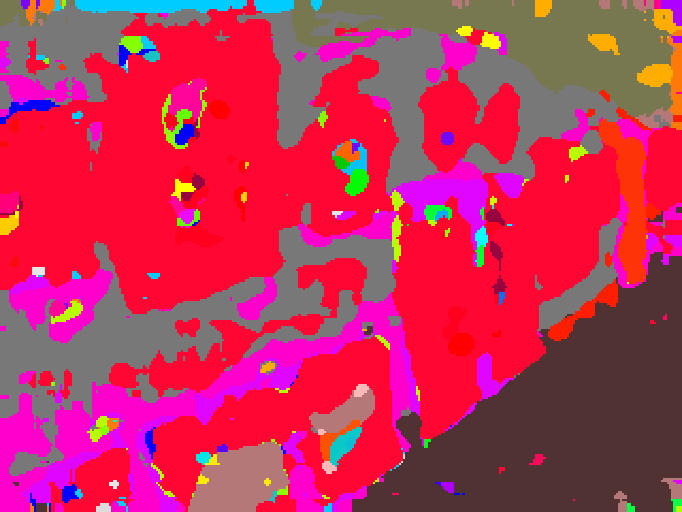} &
    \includegraphics[width=\myheight]{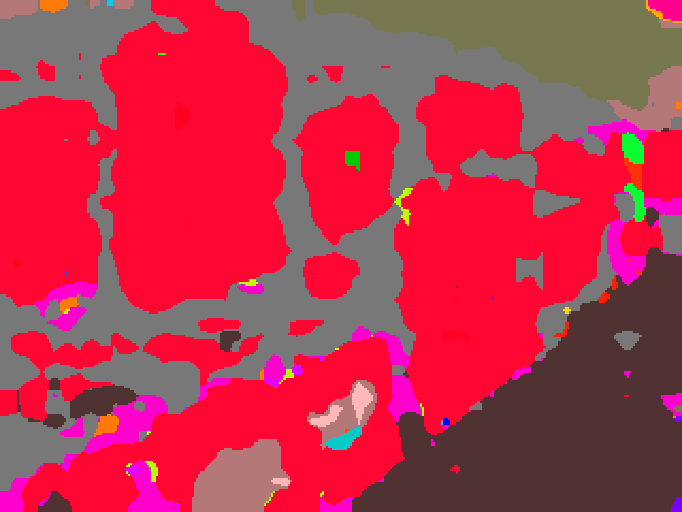}  &
    \includegraphics[width=\myheight]{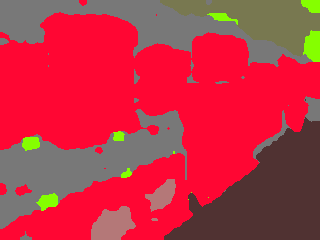} &
    \includegraphics[width=\myheight]{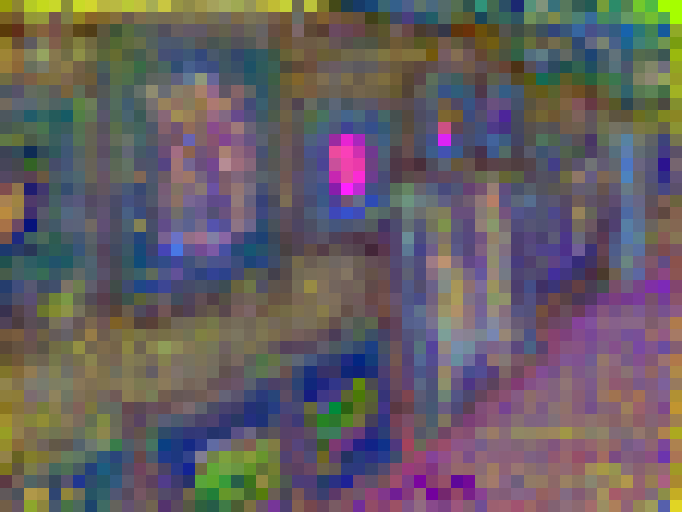} &
    \includegraphics[width=\myheight]{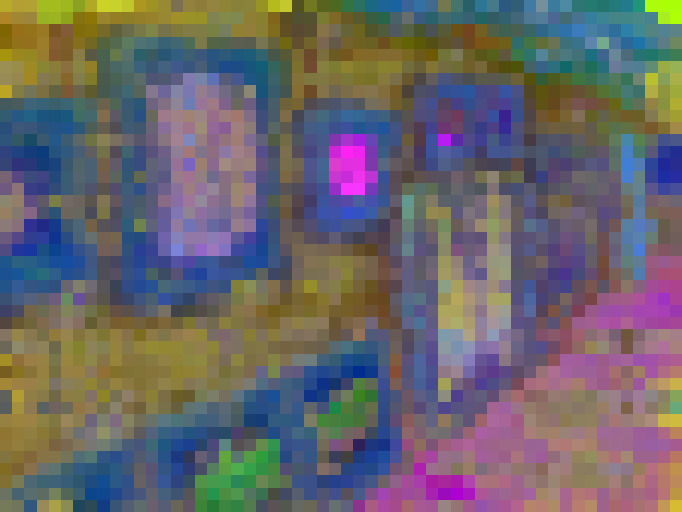} \\
    
    \includegraphics[width=\myheight]{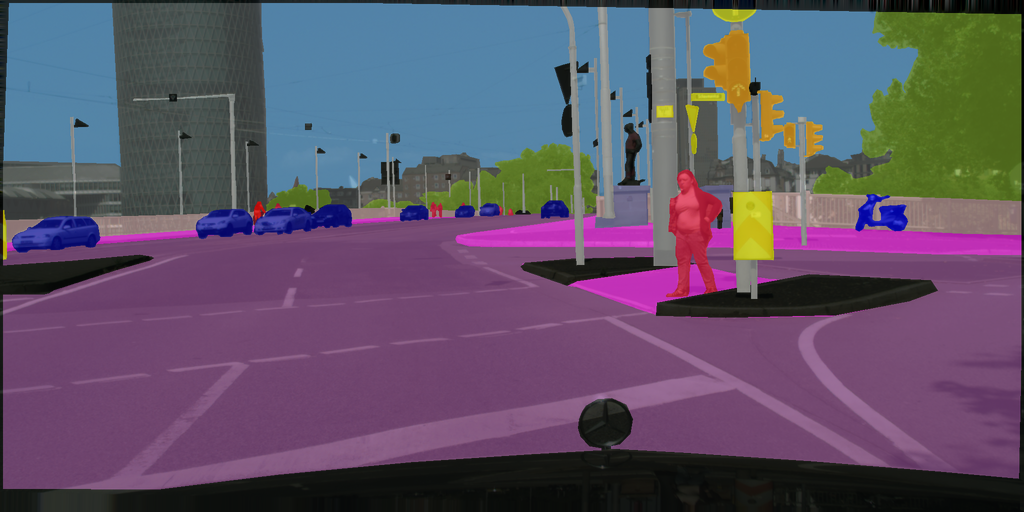} &
    \includegraphics[width=\myheight]{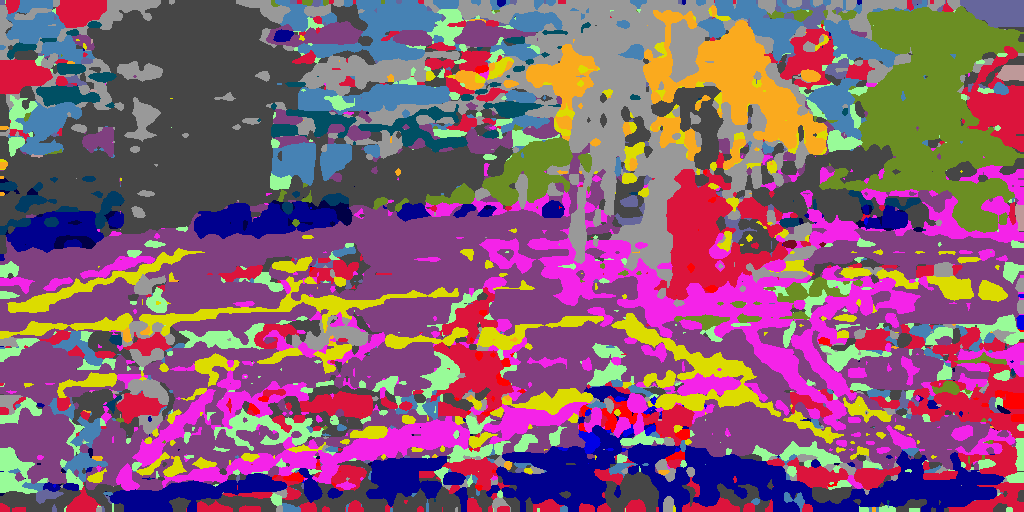} &
    \includegraphics[width=\myheight]{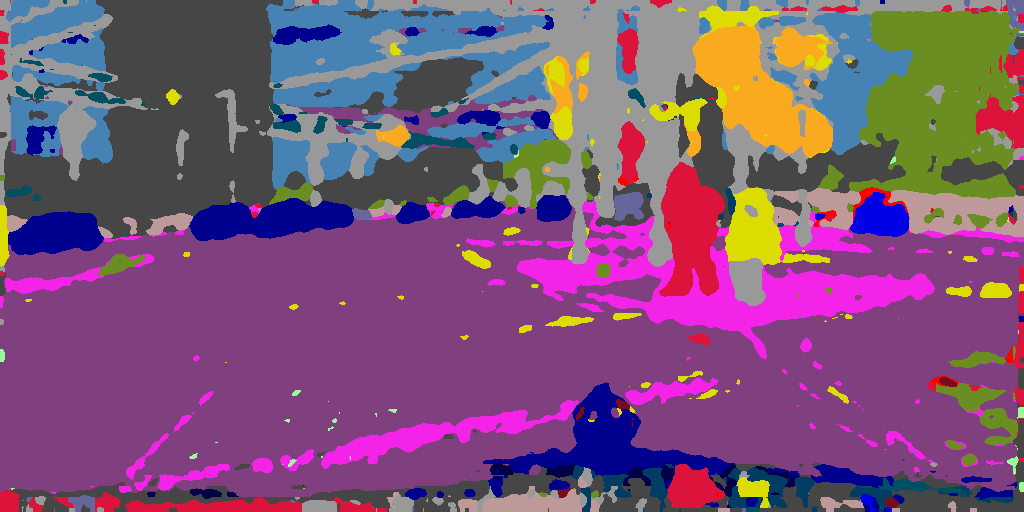} &
    \includegraphics[width=\myheight]{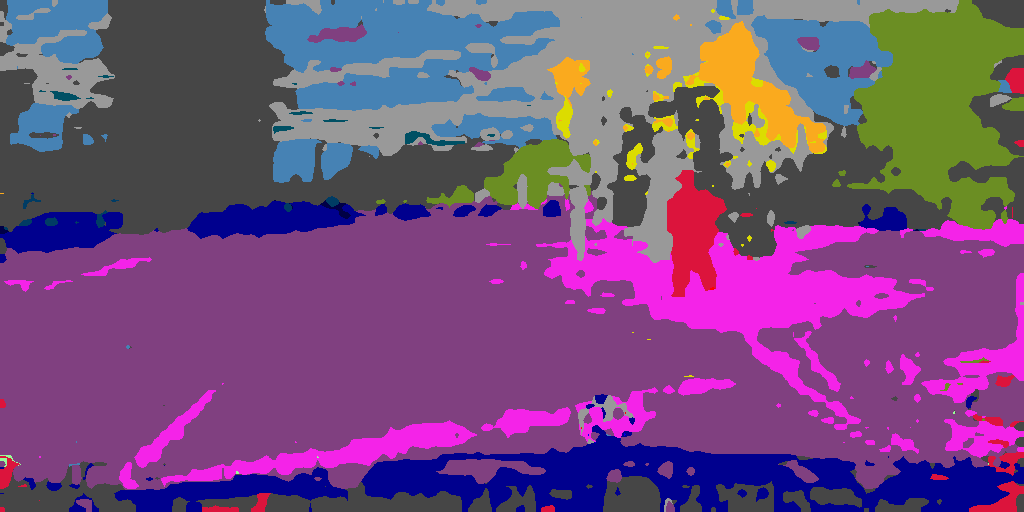} &
    \includegraphics[width=\myheight]{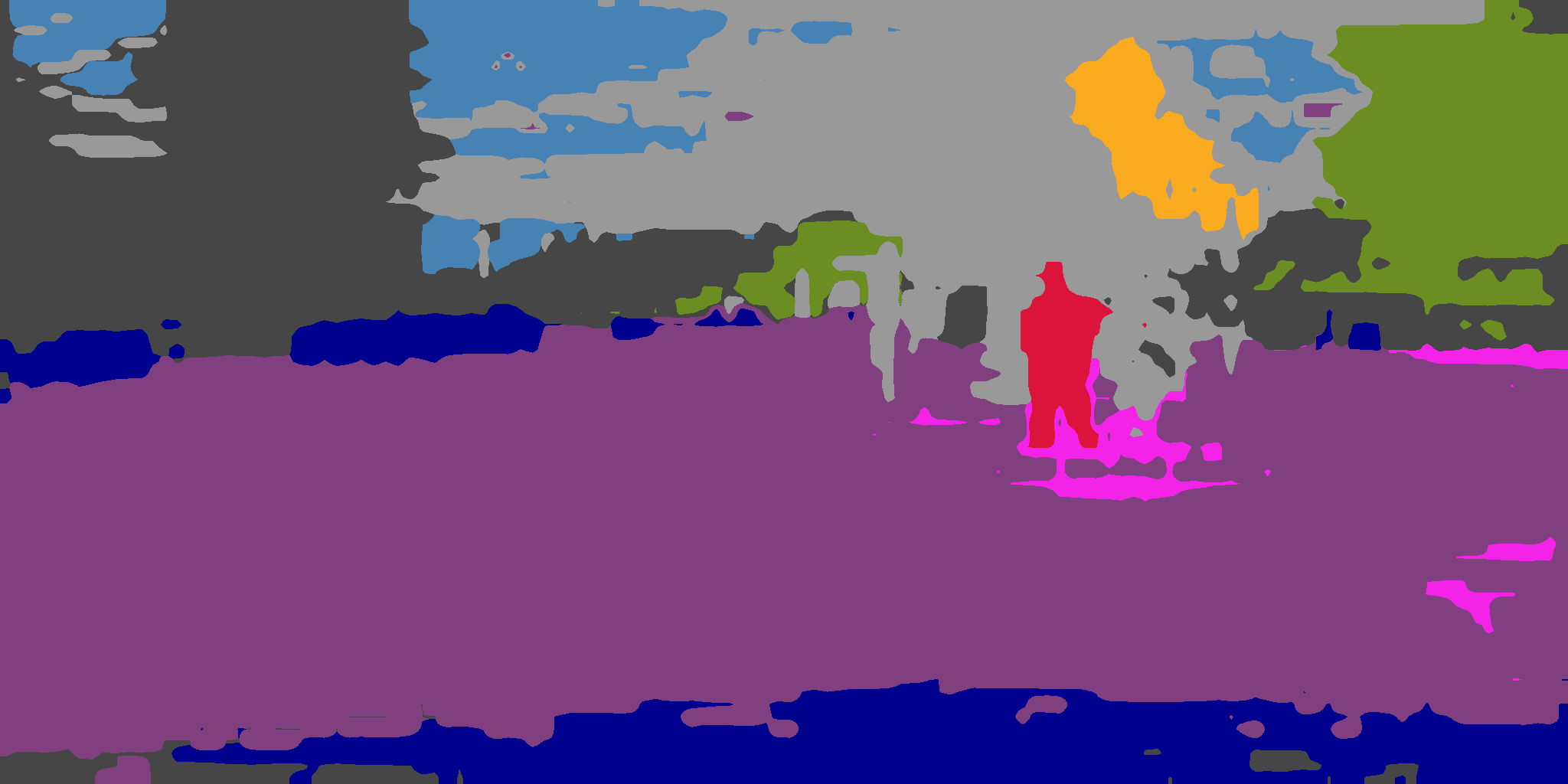} &
    \includegraphics[width=\myheight]{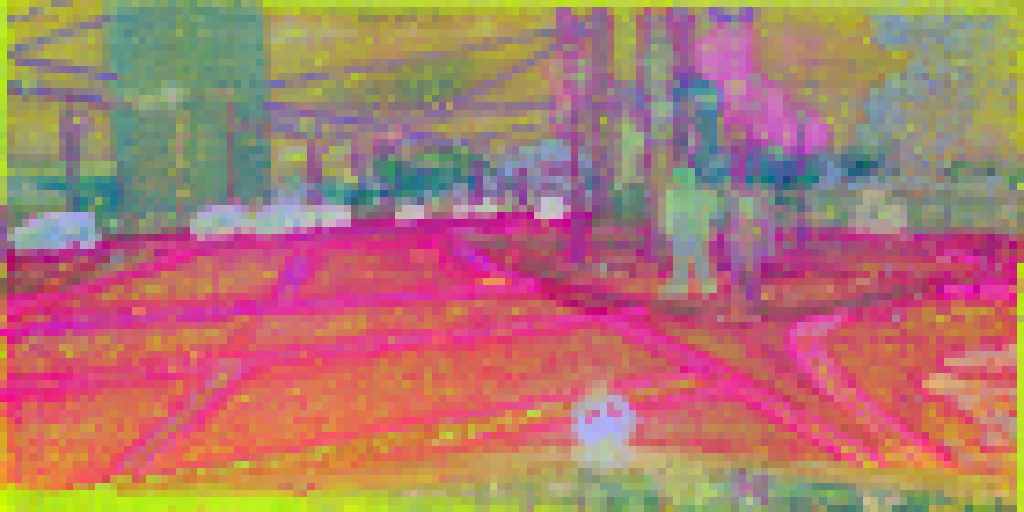} &
    \includegraphics[width=\myheight]{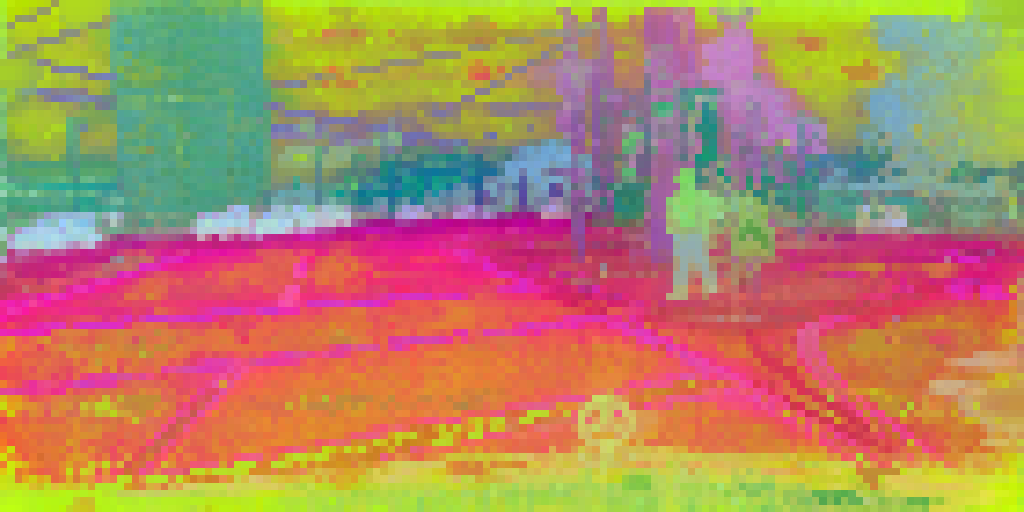} \\

\vspace{-5pt}
\end{tabular}

\vspace{-15pt}
\caption{\textbf{Qualitative segmentation and PCA analysis results.} We show 
that \ours{} yields less noisy and spatially smoother predictions than the teacher, further improved by LPOSS \citep{stojnic2025lposs}. Images are from VOC21 \citep{voc}, Context60 \citep{context}, ADE20K~\citep{ade20k} and Cityscapes~\cite{cityscapes}.}
\vspace{-10pt}
\label{fig:qual_semseg_camera_ready}
\end{figure*}

\noindent\textbf{Influence of the distillation dataset.} 
In \cref{tab:siglip2_training_set_ablation}, we validate that \ours{} is not dependent on in-domain data. When trained on the union of ADE20K, Cityscapes, and VOC training splits, instead of an SA-1B subset, the performance on corresponding validation images drops slightly (-0.2 mIoU). Restricting the data to ADE20K images preserves performance, except on Cityscapes, where larger, unseen image resolutions lead to a $\sim$1.9 mIoU drop. We initially sampled 25k SA-1B (out of 11M) images to roughly match the combined size of ADE20K, Cityscapes, and VOC training splits, and then examined whether smaller subsets suffice. Using 5\%, 10\%, and 50\% of this data, we observe consistent performance gains with larger subsets up to 25k images, beyond which additional samples yield no further improvement. When the 25k SA-1B subset is randomly resampled, results remain consistent, indicating that \ours{} is robust to the specific choice of distillation data.
\begin{table}[t]
  \vspace{5pt}    
  \centering
  \scriptsize
    \small
\renewcommand{\arraystretch}{1.1}
\begin{tabular}{lccc}

    \toprule
      SigLIP2 -- \texttt{ViT-B-16}  & VOC21 & CS  & ADE\\ 
    \midrule
    \rowcolor{gray!10} \multicolumn{4}{c}{Linear Probe}\\
    Pre-trained single-pass  & 67.1 & 54.1 & 36.0   \\
    \ours{} (Last 2 blocks) & \textbf{70.2} & 56.2 & \textbf{38.1}  \\
    \ours{} (ALL) & 68.9 &\textbf{ 66.7} & 36.5 \\
    \rowcolor{gray!10} \multicolumn{4}{c}{Hummingbird} \\
    Pre-trained single-pass  & 31.9 & 26.7 & 13.5 \\
    \ours{} (Last 2 blocks) & 41.0 & 30.0 & 16.0  \\
    \ours{} (ALL) &\textbf{54.4} & \textbf{39.9} & \textbf{22.7}  \\
    \rowcolor{gray!10} \multicolumn{4}{c}{Mask-Oracle Panoptic Segmentation}\\
    Pre-trained single-pass & N/A & 31.5 & 33.6  \\
    \ours{} (Last 2 blocks) &N/A &  28.6 & 34.0 \\
    \ours{} (ALL) & N/A &  \textbf{52.4} & \textbf{41.1} \\

    \bottomrule
    
  \end{tabular}

  \vspace{5pt}    
  \caption{\textbf{\ours{} on vision-only dense prediction tasks}. We show mIoU for linear probing and Hummingbird~\citep{hummingbird} segmentation on VOC21, Cityscapes (CS) and ADE20K. Oracle panoptic segmentation~\citep{martinovic2025dearli,vsaric2025holds} shows PQ on the latter two datasets. Improvement across all benchmarks attests to \ours{}'s benefit to visual representation quality. Best result is \textbf{bolded}.}
  \label{tab:visual_tasks}
  \vspace{-0pt}
\end{table}

\noindent\textbf{\ours{} on vision-only  tasks.}
Our contribution with \ours{} has applications beyond OVS, as it operates on the vision encoder only. 
To demonstrate this, we evaluate on vision-only dense prediction tasks (see \cref{tab:visual_tasks}), including linear probing, kNN segmentation (Hummingbird~\citep{hummingbird}), and oracle-guided panoptic segmentation via mask pooling~\citep{martinovic2025dearli, vsaric2025holds}. \ours{} consistently improves performance in all settings, with the largest gains on Cityscapes, whose resolution deviates most from pretraining. Full network fine-tuning (ALL) further enhances performance. Additional details are provided in the supplementary material.

\subsection{Qualitative Analysis}
\Cref{fig:qual_semseg_camera_ready} presents qualitative results on VOC21, Context60, ADE20K, and Cityscapes for \ours{} with SigLIP2. \ours{} not only matches, but often surpasses the teacher in spatial smoothness and semantic coherence, producing cleaner boundaries and reduced noise (e.g., the paintings example). PCA visualizations further highlight improved feature structure, with increased inter-object separability and smoother intra-object consistency without loss of fine details. For instance, the cat is more clearly separated from the cushion, while street scenes exhibit sharper transitions that preserve structure. As expected from distillation, errors largely mirror those of the teacher, including misclassifications and noisy regions, though these can be mitigated with complementary methods such as LPOSS. Additional examples are provided in the supplementary material.

\section{Conclusion}
\label{sec:conclusion}
We improve ViTs for dense feature extraction beyond their fixed pre-training resolution. Our method, \ours{}, distills from a costly sliding-window teacher into a single-pass model that tolerates varying image resolutions while being significantly faster. With no additional labeled data or architectural changes, \ours{} achieves performance comparable to, or even surpassing, the teacher on open-vocabulary semantic segmentation; gaining up to 10.5 mIoU over the single-pass baseline and reducing inference time by up to $52\times$. We demonstrate the effectiveness of our lightweight post-training technique with three different ViTs: SigLIP2, OpenCLIP and DINOv3.txt.
\ours{} is additionally shown to be effective in vision-only dense predictions tasks, with the trained models likely to be found useful for applications beyond the ones examined in this work. 

\textbf{Acknowledgements} This work was supported by the Junior Star GACR GM 21-28830M,   Croatian Recovery and Resilience Fund -- NextGenerationEU (grant C1.4 R5-I2.01.0001) and Croatian Science Foundation (grant IP-2020-02-5851). The access to the computational infrastructure of the OP VVV funded project CZ.02.1.01/0.0/0.0/16\_019/0000765 ``Research Center for Informatics'' is also gratefully acknowledged. We thank Tomáš Vojíř, Ivan Grubišić and Noa Garcia, for valuable feedback that helped improve the manuscript.
In memory of Siniša Segvić, who passed away before this work was published. He will be missed as a dear colleague, mentor, friend, deep thinker, and moral compass.

{
    \small
    \bibliographystyle{ieeenat_fullname}
    \bibliography{main}
}

\clearpage
\setcounter{page}{1}
\maketitlesupplementary

\section{Training Details}
\label{sec:training_details}

We conduct all training on two \texttt{NVIDIA RTX~A6000} GPUs using precomputed image features generated by the corresponding teacher model in sliding-window mode with a stride of 24. 
For feature generation, we use the smallest upsampling factor $r$ such that the upsampled feature maps of cropped windows will have their corresponding image sub-patches fully overlapping. Stitching is then done by simply aligning and averaging feature maps. 
For $s=24, P=16$, we choose $r=2$ which is the smallest value of $r$ for which $s$ is divisible by $P/r$.
Both the up- and downsampling is bilinear.
To accommodate variable input sizes despite using precomputed features, we train with a batch size of 1. Larger batches would require generating teacher features in grouped batches so that all samples share the same sequence length. This would either reduce the effective diversity of the training set, as image tuples would always co-occur in the same context, or introduce additional complexity to gradient accumulation, along with masking in attention to prevent cross-sample interaction. Attention does not usually support unpadded sequences of different lengths in a batch, which would necessitate flattening together all sequences into a single long one. Training is performed in mixed \texttt{float16} precision.

By default, for the initial rescaling in image augmentation, we use MMCV's {\small \texttt{RandomResize}} with  {\small \texttt{scale=(2048,1024)}}, {\small \texttt{ratio\_range=(0.5,1)}} and {\small \texttt{keep\_ratio=True}}. For the extended image range experiments, denoted by $\dagger$, we adjust {\small \texttt{scale=(2560,2560)}} and {\small \texttt{ratio\_range=(0.2,1)}}, while still keeping aspect ratio intact with {\small \texttt{keep\_ratio=True}}.
The dataset class names and their feature extraction are as in \citep{SCLIP} and its official Github implementation.
The costs for \ours{} SigLIP2 -- \texttt{ViT-B-16} time-wise include 9 hours for feature extraction on a single A6000 GPU, and 1.5 hours of training on 2 A6000. Feature storage takes about 170GB.

For experiments involving SA-1B~\citep{sam}, we use the first 25k images from
archive files \texttt{sa\_000000.tar}, \texttt{sa\_000001.tar}, and \texttt{sa\_000002.tar}, as named in the master file for downloading SA-1B.  For the alternate seed used in \cref{tab:siglip2_training_set_ablation}, we instead sample images sequentially from the randomly chosen \texttt{sa\_000165.tar}, \texttt{sa\_000205.tar}, and \texttt{sa\_000569.tar}.

\section{Performance Across Seeds}
\label{sec:repeatability}

We report repeated experimental trials to assess the repeatability of \ours{}. For each setting, we compute the mean and standard deviation across three independent runs. The aggregated results for individual tuning configurations experiments are presented in \cref{tab:config_ablations_stride_24_teacher_mean_std,tab:siglip2_training_set_ablation_mean_std}, and mirror \cref{tab:config_ablations_stride_24_teacher,tab:siglip2_training_set_ablation} in the main paper. By $\text{Mean}_6$ we denote the average mIoU over six datasets: Voc21, Voc20, Cityscapes, ADE20K, Context60, and Context59.  
$\overline{\text{Mean}_6}$  and $\sigma_{\text{Mean}_6}$ indicate the mean and standard deviation of $\text{Mean}_6$ over three independent runs.

\begin{table}[t]
  \centering
  \footnotesize
    \setlength{\tabcolsep}{3.5pt}
\begin{tabular}{lcc}
\toprule
     Training Configuration & $\mathrm{Reported\: Mean}_6$ & $\overline{\mathrm{Mean}_6}$ $\pm$ $\sigma_{\mathrm{Mean}_6}$ \\
     \midrule
   
    \rowcolor{gray!10} \multicolumn{3}{c}{\ours{} model}\\
     All params &  42.5	&42.3	$\pm$ 0.27\\ 
     Last block & 43.3	&43.2	$\pm$ 0.14 \\ 
     \textbf{Last 2 blocks (default)}   &\textbf{43.6}	&\textbf{43.6	$\pm$ 0.04} \\
     Last 3 blocks& 42.9	&43.1	$\pm$ 0.15\\ 
     Patch projection & 38.0	&38.1	$\pm$ 0.11 \\ 
     Positional encoding &39.6	&39.6	$\pm$ 0.03  \\ 
     Last 2 blocks - \texttt{MLP} & 42.7	&42.8	$\pm$ 0.07 \\ 
     Last 2 blocks - \texttt{QKV}  &41.7	&41.7	$\pm$ 0.04 \\ 
    \bottomrule
\end{tabular}

  \caption{\textbf{Effect of fine-tuning different network elements (mean and std performance across three seeds)}. We show mean mIoU over six datasets, along with the mean and standard deviation across three independent runs, for various training configurations of \ours{}-SigLIP2 -- \texttt{ViT-B-16}. Fine-tuning only the last two transformer blocks, \textbf{bolded} for emphasis, achieves the best overall performance and maintains the second smallest standard deviation.}
  \label{tab:config_ablations_stride_24_teacher_mean_std}
\end{table}

\begin{table}[t]
  \centering
  \footnotesize
    \begin{tabular}{lcc}
\toprule
    
     Training Set & $\mathrm{Reported \: Mean}_6$ & $\overline{\mathrm{Mean}_6}$ $\pm$ $\sigma_{\mathrm{Mean}_6}$ \\
    \midrule

    \rowcolor{gray!10}  \multicolumn{3}{c}{\ours{} model}\\
     ADE20k+CS+VOC  &43.4	&43.6	$\pm$ 0.27\\
     ADE20k & 43.0	&43.1	$\pm$ 0.12 \\[5pt]
     SA-1B ~1.25k (5\%)  & 41.1	&41.0	$\pm$ 0.09 \\ 
     SA-1B ~2.5k (10\%) & 42.1	&41.9	$\pm$ 0.31 \\ 
     SA-1B 12.5k (50\%) & 43.2	&43.0	$\pm$ 0.15 \\ 
     SA-1B ~~25k  (100\%)&43.6	&43.6	$\pm$ 0.04\\
     SA-1B ~~25k (diff. subset)  & 43.6	&43.3	$\pm$ 0.30\\ 
     SA-1B ~~50k  (200\%)& 43.6	&43.5	$\pm$ 0.10\\
    
    \bottomrule
\end{tabular}

  \caption{\textbf{Impact of distillation data composition and scale (mean and std performance across three seeds)}. We show mean mIoU over six datasets, along with the mean and standard deviation across three independent runs, for \ours{}-SigLIP2 -- \texttt{ViT-B-16} when training on different unlabeled sources. Comparable results across all settings show \ours{} does not rely on in-domain data, and performance saturates at 25k distillation images.}
  \label{tab:siglip2_training_set_ablation_mean_std}
\end{table}

\section{Measuring Inference Time}
\label{sec:timing_details}

In this section, we provide details on how we measure inference time for the experiments reported in \cref{fig:stride_vs_performance}. To ensure a fair comparison, we only accumulate the time required for the forward passes needed to process each image. In single-pass mode, this corresponds to timing the underlying Vision Transformer (ViT) for a single image. For sliding-window inference, we sum the time required to process each sub-batch of window crops. The sub-batch size is set to 60 and is kept constant across experiments; if an image's total number of windows is smaller, they are processed in a single batch. All experiments are conducted on a single \texttt{NVIDIA RTX~A6000} GPU, and inference time is measured by accumulating the differences between start and end timestamps using the native \texttt{time} package. Before timing each forward pass, we perform 10 warm-up passes with the data. Measurement is done on the 11th pass.

\section{SPAR with LPOSS}
\label{sec:spar_with_lposs}

For experiments combining \ours{}-SigLIP2 with LPOSS \citep{stojnic2025lposs}, we tune the $\gamma$ hyperparameter, which controls the Laplacian computation during label propagation. For single-pass inference, we set $\gamma=10$ to better align with the label distribution produced by \ours{}, while for pretrained single-pass and sliding-window inference we retain the default $\gamma=1$, as we found these settings to perform best for each approach.

\section{Vision-only dense prediction tasks details}
\label{sec:dense_prediction_tasks_details}

We utilize the code of \citep{benchmarking-benchmark} with only minor adaptations to enable training and evaluation in a single pass using native image resolutions. We disable scale jittering and cropping augmentations in the training source code: scale jittering would distort images in a way that does not reflect reality, and cropping to a fixed size is inconsistent with our goal of training a transformer capable of processing images at their native aspect ratio. Horizontal flipping remains enabled during training.
In \cref{tab:visual_tasks_supp}, we additionally report linear probing results when using the code’s default augmentations and resizing of images: $512^2$ for VOC21 and ADE20K, and $1024^2$ for Cityscapes. Training the last two blocks and all parameters becomes more similar for VOC21 and ADE20K, while the gap on Cityscapes closes. \ours{} still provides a noticeable performance benefit.

\begin{table}[t]
  \centering
  \footnotesize
    \begin{tabular}{lccc}

    \toprule
      SigLIP2 -- \texttt{ViT-B-16}  & VOC21 & CS  & ADE\\ 
    \midrule
    \rowcolor{gray!10} \multicolumn{4}{c}{Linear Probe - native resolution}\\
    Pre-trained single-pass  & 67.1 & 54.1 & 36.0   \\
    \ours{} (Last 2 blocks) & \textbf{70.2} & 56.2 & \textbf{38.1}  \\
    \ours{} (All) & 68.9 &\textbf{ 66.7} & 36.5 \\
    
    \rowcolor{gray!10} \multicolumn{4}{c}{Linear Probe - repository resolution}\\
    Pre-trained single-pass  & 71.2	& 57.0	&37.7   \\
    \ours{} (Last 2 blocks) & 74.9	&60.9	&\textbf{40.0} \\
    \ours{} (All) &\textbf{ 75.0}	&\textbf{67.1}	&39.1 \\

    \bottomrule
    
  \end{tabular}

  \caption{\textbf{\ours{} linear probing performance using different resolutions}. We show mIoU for linear probing on VOC21, Cityscapes (CS) and ADE20K when utilizing native image resolution vs default resizing of \citep{benchmarking-benchmark} for training and evaluation. Improved results across all benchmarks show \ours{}'s benefit to visual representation quality. Best result is \textbf{bolded}.}
  \label{tab:visual_tasks_supp}
\end{table}

We use the official implementation of Hummingbird \citep{hummingbird} for KNN segmentation, leaving images at their native resolution and aspect ratio. Due to A6000 memory limitations, we utilize only 50\% of VOC21 and 30\% of ADE20K training images to construct the index used to classify patches from the evaluation images. Cityscapes uses the full training set, while the other two datasets use the largest subset that avoids out-of-memory errors.  We report the mean over three trials, as the training images are randomly sampled, and observe minimal fluctuations: the standard deviation never exceeds 0.4 mIoU points.

The panoptic segmentation experiments report ADE20K results for images resized to $800^2$, while Cityscapes is kept at its native resolution, following standard practice. We additionally evaluated other resolutions and observed the same trend: training all parameters yields the highest-quality representations and emerges as a promising approach for future research in resolution-agnostic panoptic segmentation.

\begin{table}[t]
  \centering
  \footnotesize
    \setlength{\tabcolsep}{2.8pt} 
  \begin{tabular}{lccccc}
    \toprule
       GT Panoptic & \multicolumn{4}{c}{ADE} & \multicolumn{1}{c}{CS} \\
          & $512^2$ & $640^2$ & $800^2$ & $1024^2$ & $1024 \times 2048$ (native) \\
          \midrule
         Pre-trained single pass & 33.9	&34.3	&33.6	&31.0	&31.5  \\ 
          \ours{} Last 2 blocks & 34.6	&35.2	&34.0	&30.7	&28.6\\   
          \ours{} All parameters & \textbf{37.9} & \textbf{39.7}	&\textbf{41.1}	&\textbf{39.0}	&\textbf{52.4}\\  
    \bottomrule
    \end{tabular}

  \caption{\textbf{\ours{} mask-oracle panoptic segmentation results for different resolutions}. We show PQ for panoptic segmentation on Cityscapes (CS) and ADE20K when using ground truth masks for pooling features~\cite{martinovic2025dearli,vsaric2025holds}. Improved results across  resolution show \ours{}-All's benefit to visual representation quality. Best result is \textbf{bolded}.}
  \label{tab:panoptic_segmentation_supp}
\end{table}

\section{\ours{} and other distillation schemes }
\label{sec:distillation_schemes}

In \cref{tab:distillation_schemes}, we explore additional distillation targets. We isolate the effect of multi-resolution training by distilling from a single-pass teacher, for which we precompute the image features and maintain the same setup as described in \cref{sec:experiments}. During training, the student sees the image additionally bilinearly interpolated by a random factor and attempts to align its features, which are bilinearly up- or downsampled to match the teacher’s as needed. As observed, this yields only a +2.9 mIoU average improvement, highlighting the importance of the teacher’s multi-context supervision. We also explore using a teacher with a finer stride of $s=16$, which underperforms by 3 mIoU, emphasizing the benefit of observing pixels in the context of different patches.

To quantify potential benefits that might be missed by not aligning class similarities, we experiment with using the class lists of ADE20K and Cityscapes, respectively. This approach overfits to the domain of the class set used (e.g., using Cityscapes classes yields high performance on Cityscapes but not on other datasets), reaffirming that \ours{} does not require any knowledge of the target domain to be effective and in fact benefits from the classless approach.

\begin{table}[t]
  \centering
  \footnotesize
    \setlength{\tabcolsep}{3pt} 
  \begin{tabular}{lllccccc}
        \toprule
             SigLIP2 -- \texttt{ViT-B-16}    & Voc21 & Voc20 & CS & ADE & C60 & C59 & $\text{Mean}_{6}$ \\ 
            \midrule 
             Pre-trained single-pass       & 36.1	&71.3 &23.5 &16.8 &24.5 &26.1 &33.1 \\ 
             Multi-resolution distill.          & 38.0	&76.8	&26.3	&19.0	&26.8	&29.0	&36.0\\
             \ours{} w/ Teach. $s=16$       & 43.8	&77.6	&36.2	&21.3	&31.0	&34.0	&40.6 \\
             \ours{} w/ Teach. $s=24$        & 47.3 & \textbf{81.5	}&\textbf{38.4} & \textbf{23.4}	& \textbf{33.8} & \textbf{37.2}	& \textbf{43.6}  \\
             Class sim. distill. ADE          & \textbf{48.0}	 &75.4	 &37.1	 &23.0	 &32.8	 &36.4	 &42.1 \\
             Class sim. distill. CS        & 44.0	 &64.8	 &38.3	 &19.8	 &26.0	 &31.8	 &37.4 \\
            
       \bottomrule
        \end{tabular}

  \caption{\textbf{\ours{} compared to other distillation schemes}. We show mIoU over six datasets, along with the mean. The results underscore the importance of \ours{} using a sliding-window teacher with a stride allowing for non-overlapping patches between windows as well as aligning at the visual feature level. Distilling class similarities overfits to the utilized classes. Best results are \textbf{bolded}.}
  \label{tab:distillation_schemes}
\end{table}

\section{Qualitative Analysis}
\label{sec:additional_qualitative_analysis}

\begin{figure*}[t]
\vspace{-8pt}
\centering
\setlength{\tabcolsep}{1pt}
\renewcommand{\arraystretch}{0.8}
\input{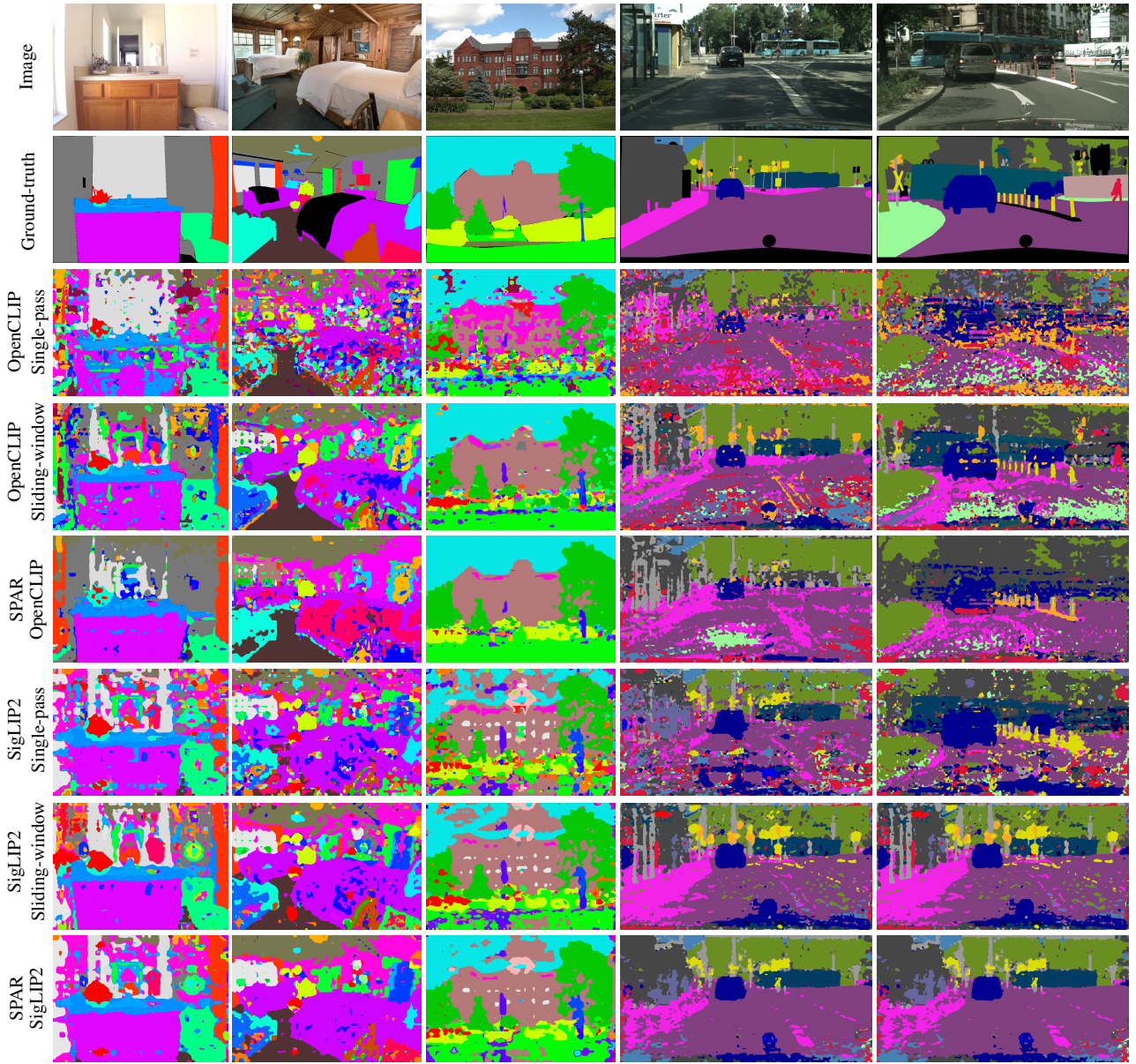}
\vspace{-8pt}
\caption{\textbf{Qualitative segmentation results.} We show 
\ours{} yields more coherent and smoother predictions than the teacher. Images are from ADE20K~\citep{ade20k} (first three columns) and Cityscapes~\cite{cityscapes} (last two columns). }
\label{fig:semseg_supp}
\end{figure*}

\begin{figure*}[t]
\vspace{-8pt}
\centering
\setlength{\tabcolsep}{1pt}
\renewcommand{\arraystretch}{0.8}
\input{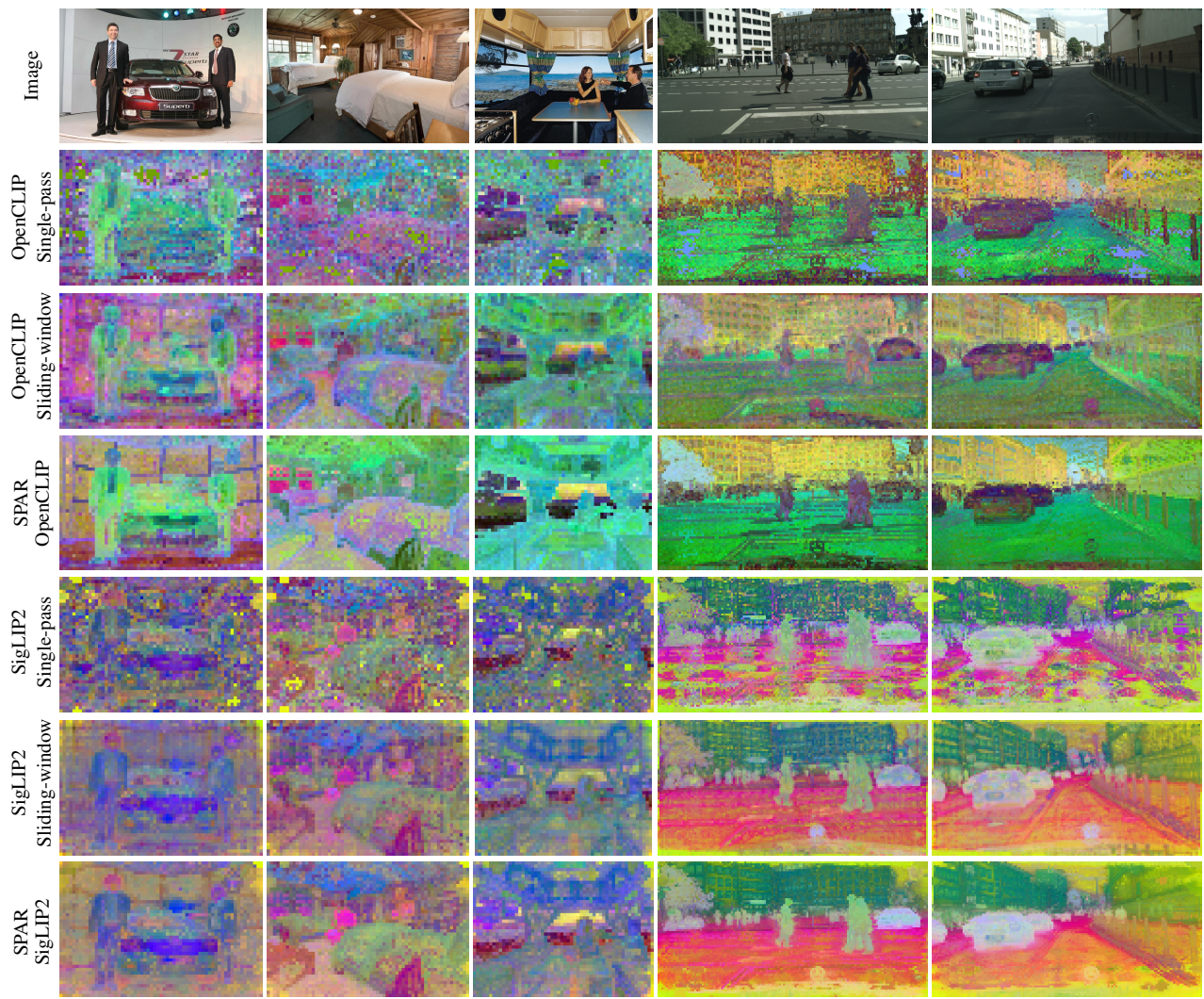}
\vspace{-8pt}
\caption{\textbf{Feature quality visualization via PCA.} We project features onto the same PCA basis, computed from the teacher’s sliding-window representations, to enable a consistent and semantically aligned comparison across models.  \ours{} improves inter-object separability and smooths intra-object consistency without losing fine details. The former is visible in the clearer distinction between people and the wall behind them (1st column) or pedestrians on the street (4th column), while improved intra-object consistency is most apparent in the interior of the camper (3rd column). Images are from ADE20K~\citep{ade20k} (first three columns) and Cityscapes~\cite{cityscapes} (last two columns).}
\label{fig:pca_supp}
\end{figure*}

We provide additional semantic segmentation results in \cref{fig:semseg_supp} and PCA visualizations in \cref{fig:pca_supp} for ADE20K~\citep{ade20k} and Cityscapes~\citep{cityscapes}. Utilized backbones, MaskCLIP~\citep{MaskCLIP} with OpenCLIP \citep{openclip} and SigLIP2~\citep{SigLIP2}, are indicated per-row in the figures. The segmentation maps demonstrate how \ours{} denoises teacher predictions while preserving semantic alignment. \ours{}-OpenCLIP improves delineation of bathroom elements and buildings (1st and 3rd columns in \cref{fig:semseg_supp}) while suppressing noisy classes on the roads (4th and 5th columns). \ours{}-SigLIP2 behaves similarly, yet slightly more robustly. 
The PCA visualizations show further how \ours{} improves inter-object separability and smooths intra-object consistency without losing finer details. The former is visible in the PCA from both backbones, e.g., in the clearer separation of people from the wall or bedroom elements (1st and 2nd columns in \cref{fig:pca_supp}), while improved intra-object consistency is most apparent in the interior of the camper (3rd column).

\end{document}